\documentclass[letterpaper, 10 pt, journal, twoside, dvipsnames]{IEEEtran}
\IEEEoverridecommandlockouts

\newif\ifarxiv
\arxivtrue

\usepackage[utf8]{inputenc}
\usepackage{graphicx} %
\usepackage{amsmath} %
\usepackage{amssymb}  %
\usepackage{tikz}
\usepackage{url}
\usepackage{xspace}
\usepackage{enumitem}
\usepackage{multirow}
\usepackage{array}
\usepackage{tabularx}
\usepackage{textcomp}
\usepackage{mathtools}
\usepackage{bm}
\usepackage{adjustbox}
\usepackage{microtype}
\usepackage[percent]{overpic}
\usepackage{nidanfloat}

\ifarxiv
\usepackage{xcolor}
\usepackage{svg}
\fi

  \usepackage{cite}

\ifarxiv
\usepackage[font=footnotesize]{caption}
\fi

\makeatletter
\let\NAT@parse\undefined
\makeatother

\ifarxiv
\usepackage{eso-pic}
\usepackage{hyperref}

\hypersetup{
  linkcolor  = NavyBlue,
  citecolor  = NavyBlue,
  urlcolor   = NavyBlue,
  colorlinks = true,
}

\usepackage{scalerel}
\usepackage{tikz}
\usetikzlibrary{svg.path}

\definecolor{orcidlogocol}{HTML}{A6CE39}
\tikzset{
  orcidlogo/.pic={
    \fill[orcidlogocol] svg{M256,128c0,70.7-57.3,128-128,128C57.3,256,0,198.7,0,128C0,57.3,57.3,0,128,0C198.7,0,256,57.3,256,128z};
    \fill[white] svg{M86.3,186.2H70.9V79.1h15.4v48.4V186.2z}
                 svg{M108.9,79.1h41.6c39.6,0,57,28.3,57,53.6c0,27.5-21.5,53.6-56.8,53.6h-41.8V79.1z M124.3,172.4h24.5c34.9,0,42.9-26.5,42.9-39.7c0-21.5-13.7-39.7-43.7-39.7h-23.7V172.4z}
                 svg{M88.7,56.8c0,5.5-4.5,10.1-10.1,10.1c-5.6,0-10.1-4.6-10.1-10.1c0-5.6,4.5-10.1,10.1-10.1C84.2,46.7,88.7,51.3,88.7,56.8z};
  }
}

\newcommand\orcidicon[1]{\href{https://orcid.org/#1}{\mbox{\scalerel*{
\begin{tikzpicture}[yscale=-1,transform shape]
\pic{orcidlogo};
\end{tikzpicture}
}{|}}}}

\fi

\newcommand{\argmin}[1]{\underset{#1}{\operatorname{arg}\,\operatorname{min}}\;}

\newcolumntype{C}[1]{>{\centering\arraybackslash}m{#1}}
\newcolumntype{Y}{>{\centering\arraybackslash}X}

\graphicspath{{figs/}}

\title{Event-Based Visual Place Recognition\\With Ensembles of Temporal Windows}

\author{%
\ifarxiv
Tobias Fischer\textsuperscript{\,\orcidicon{0000-0003-2183-017X}},~\IEEEmembership{Member,~IEEE} and Michael Milford\textsuperscript{\,\orcidicon{0000-0002-5162-1793}},~\IEEEmembership{Senior Member,~IEEE}%
\else
Tobias Fischer$^{1}$ and Michael Milford$^{1}$%
\fi
\thanks{Manuscript received: May 21, 2020; Revised September 1, 2020; Accepted September 4, 2020.}%
\thanks{This paper was recommended for publication by Editor S. Behnke upon evaluation of the Associate Editor and Reviewers' comments. This work received funding from the Australian Government, via grant AUSMURIB000001 associated with ONR MURI grant N00014-19-1-2571. The authors acknowledge continued support from the Queensland University of Technology (QUT) through the Centre for Robotics.}
\thanks{%
\ifarxiv
\else
$^{1}$%
\fi
Tobias Fischer and Michael Milford are with the QUT Centre for Robotics, Queensland University of Technology, Brisbane, QLD 4000, Australia {\tt\footnotesize tobias.fischer@qut.edu.au}}%
\thanks{This letter has supplementary downloadable material available at https://\allowbreak ieeexplore.ieee.org}
\thanks{Digital Object Identifier (DOI): \href{http://doi.org/10.1109/LRA.2020.3025505}{10.1109/LRA.2020.3025505}}%
}

\newcommand{\datasetname}{Brisbane-Event-VPR\xspace}
\newcommand{\methodname}{Ensemble-Event-VPR\xspace}
\newcommand{\ddddataset}{\mbox{DDD-17}\xspace}

\newcommand{\numtraverses}{6\xspace}

\newcommand{\performancemeanddd}{73.6}

\newcommand{\performanceimprovementdddsetone}{24.0}
\newcommand{\performanceimprovementdddsetonestd}{9.0}

\newcommand{\performanceimprovementdddsettwo}{3.7}
\newcommand{\performanceimprovementdddsettwostd}{0.1}

\newcommand{\performanceimprovementoverbestdddsetone}{13.2}
\newcommand{\performanceimprovementoverbestdddsettwo}{2.4}

\newcommand{\performanceimprovement}{43.3}
\newcommand{\performanceimprovementstd}{12.9}

\newcommand{\performanceimprovementoverbest}{34.9}
\newcommand{\performanceimprovementoverbeststd}{11.3}

\newcommand{\performanceworsemaxvote}{6.0}
\newcommand{\performanceworsemaxvotestd}{4.9}

\newcommand{\performanceworseapproximate}{9.5}
\newcommand{\performanceworseapproximatestd}{5.3}

\newcommand{\performanceimprovementapproximatebest}{21.4}
\newcommand{\performanceimprovementapproximatebeststd}{3.3}

\newcommand{\performancemean}{51.3}

\newcommand{\performanceimprovementmodel}{17.8}
\newcommand{\performanceimprovementmodelstd}{6.4}

\newcommand{\datasetlengthminutes}{66}

\begin{document}
\bstctlcite{MyBSTcontrol}

\ifarxiv
\AddToShipoutPicture*{%
     \AtTextUpperLeft{%
         \put(-3.5,10){
           \begin{minipage}{\textwidth}
              \scriptsize
              \MakeUppercase{Preprint version; final version available at} \url{http://doi.org/10.1109/LRA.2020.3025505}
           \end{minipage}}%
     }%
}
\fi

\maketitle

\begin{abstract}
Event cameras are bio-inspired sensors capable of providing a continuous stream of events with low latency and high dynamic range.
As a single event only carries limited information about the brightness change at a particular pixel, events are commonly accumulated into spatio-temporal windows for further processing. 
However, the optimal window length varies depending on the scene, camera motion, the task being performed, and other factors. 
In this research, we develop a novel ensemble-based scheme for combining temporal windows of varying lengths that are processed in parallel. For applications where the increased computational requirements of this approach are not practical, we also introduce a new ``approximate'' ensemble scheme that achieves significant computational efficiencies without unduly compromising the original performance gains provided by the ensemble approach. We demonstrate our ensemble scheme on the visual place recognition (VPR) task, introducing a new \datasetname dataset with annotated recordings captured using a DAVIS346 color event camera.
We show that our proposed ensemble scheme significantly outperforms \emph{all} the single-window baselines and conventional model-based ensembles, \emph{irrespective} of the image reconstruction and feature extraction methods used in the VPR pipeline, and evaluate which ensemble combination technique performs best.
These results demonstrate the significant benefits of ensemble schemes for event camera processing in the VPR domain and may have relevance to other related processes, including feature tracking, visual-inertial odometry, and steering prediction in driving.

\end{abstract}

\section{Introduction}
\IEEEPARstart{E}{vent} cameras output asynchronous per-pixel brightness changes (i.e.~events), rather than fixed-rate images like traditional frame-based cameras. An event is defined by its pixel coordinates, a timestamp with sub-millisecond precision, and the polarity denoting whether the brightness increased or decreased. Event cameras have recently gained popularity due to some of their advantageous properties over frame-based cameras, such as significantly higher dynamic range, very low latencies, and lack of motion blur~\cite{Gallego2019,Li2015,DAVIS346}. 

\begin{figure*}[b]
  \centering
  \includegraphics[width=0.98\textwidth]{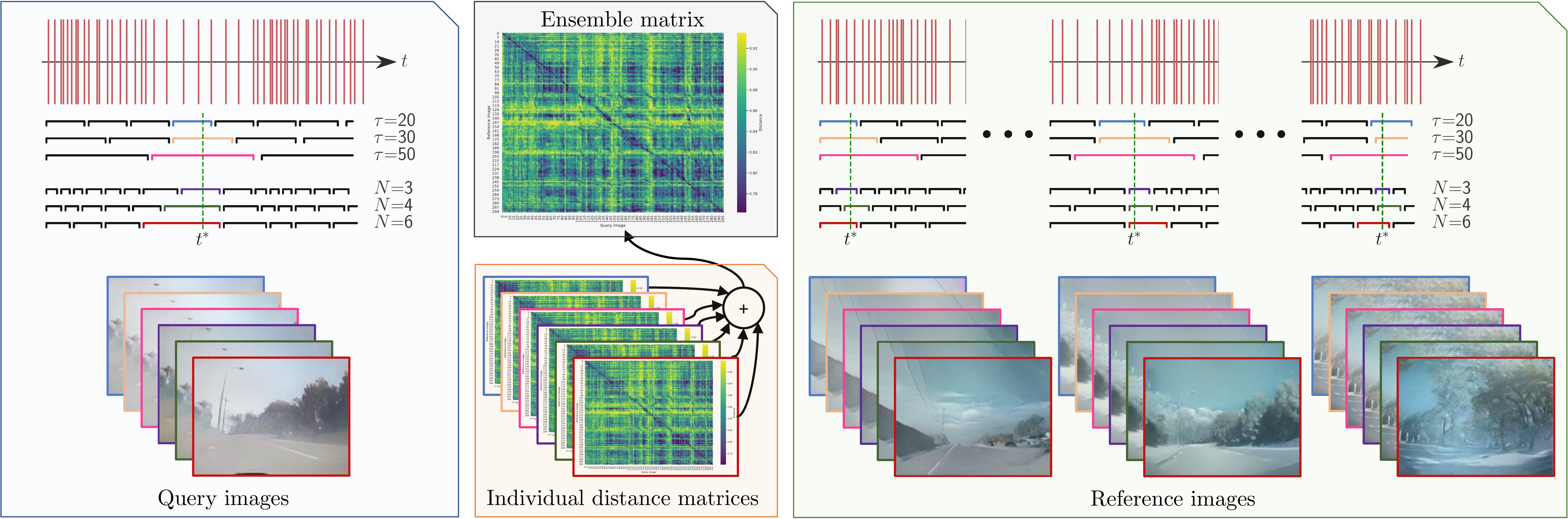}
  \ifarxiv
  \else
  \vspace*{-0.2cm}
  \fi
  \caption{\methodname. Given a query event sequence, we consider a set of temporal windows with varying time spans $\tau$ or window lengths~$N$. For each of these temporal windows, we reconstruct corresponding query images. Features extracted from these query images are compared to the corresponding features (same window time span / length; correspondence is indicated using the same border color) of the reference images, resulting in individual distance matrices. The individual distance matrices are then combined using the mean rule, resulting in the final ensemble matrix.}%
  \label{fig:catchy_fig}%
\end{figure*}

In addition to these advantages, the continuous stream of events can be interpreted in multiple ways. For example, each incoming event can be used to update the system's current state~\cite{Weikersdorfer2012,Kim2014,Kim2016}, or events can be accumulated and then batch-processed~\cite{Sekikawa2019,Gallego2017,Gallego2018,Maqueda2018,Gehrig2018,Bryner2019,Rebecq2017a}. The accumulation can be performed either over a fixed time span, or over a fixed number of events -- we refer to these accumulated events as temporal windows.

In prior work, the window length (time or number of events) for batch-processing was chosen depending on the dataset~\cite{Sekikawa2019} or based on the task performance on a validation set~\cite{Maqueda2018}. Alternatively, the window length was varied based on the amount of texture present in the scene, as more texture leads to more brightness changes and hence more events~\cite{Gehrig2018,Bryner2019,Rebecq2017a}. 

In this work, we investigate the hypothesis that it is beneficial to process multiple window lengths in parallel (see illustration in Fig.~\ref{fig:catchy_fig}). Each window length is then treated as an ensemble member, and the members are combined using a combination scheme such as the mean rule. 

Ensembles can generally be thought of as noise filters while retaining information that is common to all ensemble members~\cite{Polikar2012}. An ensemble in this letter refers to the application of the same method on different interpretations (i.e.~window lengths) of the same data, rather than the more commonly used notation of model-based ensembles that combine different methods that are given the same data as input (as in e.g.~\cite{FischerECCV2018, Zambelli2016tcds}). %

We probe our proposed ensemble scheme on the visual place recognition (VPR) task and thus name our method \emph{\methodname}. VPR is an image retrieval task, where given a set of reference images and a single query image, the task is to find the reference image captured at the same place as the query image~\cite{Lowry2016}. VPR poses many challenges: the same place can appear perceptually different depending on the time of day, season, viewpoint, \textit{etc.}, while some distant places might appear very similar~\cite{Milford2012,Lowry2016}.

While VPR has received limited attention within the event-based vision community~\cite{Milford2015}, we argue that event cameras are well suited for the VPR task. Specifically, their high dynamic range and the lack of motion blur is beneficial in challenging lighting conditions and high-speed scenarios, respectively. %

Our research presented here makes several contributions:
\begin{enumerate}[wide]
    \item We introduce a novel pipeline for large-scale VPR using event stream data. We first reconstruct multiple sets of images from the event stream, where each set is derived from a distinct window length. We then extract features from the images and build (window length specific) distance matrices for each image set by comparing the associated reference image features with query image features.
    \item We propose to treat these window-specific distance matrices as ensemble members within an ensemble scheme and combine all members using arithmetic rules such as the mean rule. Combining information contained in windows with varying length leads to superior place matches. %
    \item We devise an alternative ``approximate'' ensemble scheme which relaxes the computational burden that comes with the ensemble scheme by only using a single window size at query time.
    \item We introduce and make publicly available \textit{\datasetname}, the first event-based dataset to be suitable for large-scale VPR. This new dataset contains \numtraverses traverses of an 8km long route captured at different times of the day and in varying weather conditions. %
    \item We provide extensive evaluations of \methodname on \datasetname as well as the DAVIS Driving Dataset2017 (\ddddataset) dataset~\cite{Binas2017}. We demonstrate that the ensemble significantly outperforms each of the ensemble members which are restricted to a single window length. 
    \item We present ablation studies that demonstrate that our ensemble scheme is agnostic to the image reconstruction and feature extraction methods used, and that it outperforms conventional model-based ensembles.
\end{enumerate}

To foster future research, we make the code and dataset available: \url{https://github.com/Tobias-Fischer/ensemble-event-vpr}.%

\section{Related Works}
We first review how prior works determine the window size for event batch processing (Section~\ref{subsec:rel-windowsize}). We then briefly provide an overview of related works on VPR in \mbox{Section~\ref{subsec:rel-vpr}}. %
Finally, we review related datasets captured with event cameras (Section~\ref{subsec:rel-eventdatasets}).%

\vspace*{-0.15cm}
\subsection{Window Size in Event-Based Processing}
\label{subsec:rel-windowsize}
The motivation for batch-processing of events is twofold: a single event does not carry much information, and updating the system's state with each incoming event can be computationally expensive.

Gallego and Scaramuzza~\cite{Gallego2017} build event images from a sliding window of events, and motion compensate the events within the window. The same authors have subsequently shown that maximizing the contrast of warped event images can be used for a variety of downstream tasks~\cite{Gallego2018}. In both of these works, windows of fixed length are considered, and the window length's impact is not further investigated.

An investigation of the window length can instead be found in~\cite{Gehrig2018,Bryner2019}, where the window length is adjusted based on the amount of texture in the scene. Mueggler et al.~\cite{Mueggler2015} propose to estimate the lifetime of \emph{individual} events, which could be considered as sub-regions of the image that have their own adaptive spatio-temporal window. Rebecq et al.~\cite{Rebecq2017a} proposed the adaptation of the window length based on the number of points in the current 3D map.

Other works used windows of a fixed time interval~\cite{Sekikawa2019,Maqueda2018}, rather than windows containing a fixed number of events. However, depending on the dataset, the window size significantly changes (from 32ms to 128ms). Maqueda \textit{et al.}~\cite{Maqueda2018} argue that windows with a too short interval (10ms) are not sufficiently discriminative as they do not contain enough information, while an interval too long (200ms) washes out contours of objects. We note that Gallego and Scaramuzza~\cite{Gallego2017} argue that using a fixed number of events, rather than a fixed time interval, preserves the data-driven nature of event cameras and is thus preferable.

Finally, we mention the Hierarchy Of event-based Time Surfaces (HOTS) \cite{HOTS2017}, where each level of the hierarchy extracts increasingly abstract features using spatio-temporal windows of increasing sizes. While HOTS has a hierarchical layout, we process multiple time windows in parallel instead so we can build an ensemble scheme. Ensembles have generally been shown to increase accuracy as the error components of the ensemble members are averaged out~\cite{Polikar2012}.

\vspace*{-0.15cm}
\subsection{Visual Place Recognition}
\label{subsec:rel-vpr}

Visual Place Recognition (VPR) refers to the problem of deciding whether or not a place has been visited before; and if it has been visited before, which place it was. VPR has been investigated in great detail, as VPR can, for example, be used to propose loop closure candidates for simultaneous localization and mapping (SLAM) systems. VPR is closely related to visual localization, i.e.~estimating the 6 degrees-of-freedom camera pose of a query image with respect to a reference image~\cite{Sattler_2018_CVPR}. As an excellent survey on VPR can be found in~\cite{Lowry2016}, we limit our review to SeqSLAM~\cite{Milford2012} and NetVLAD~\cite{Arandjelovic2018}, which are used in Section~\ref{subsec:vpr} to extract features, as well as prior work on VPR using event cameras.

Sequence SLAM (SeqSLAM)~\cite{Milford2012} is one of the most popular VPR methods that rely on hand-crafted features. The key contribution was in dividing place matching into two steps: In the first step, place matching is performed for short, local navigation sequences rather than across the whole dataset. The second step is then to recognize coherent sequences of these ``local best matches''. %

NetVLAD~\cite{Arandjelovic2018} (VLAD for Vector of Locally Aggregated Descriptors) is a popular learning-based approach for VPR. The underlying convolutional neural network is remarkably robust, even when applied to data domains that largely differ from the training dataset. The authors use Google StreetView data to train their network by employing a weakly supervised triplet ranking loss.

Little previous work has focused on VPR using event cameras. In the crude approach by~\cite{Milford2015}, the events were simply accumulated over a 10ms time window and then down-sampled to low-resolution images input into SeqSLAM~\cite{Milford2012}.

\vspace*{-0.25cm}
\subsection{Event Datasets}
\label{subsec:rel-eventdatasets}
There is a wide range of datasets captured using event cameras, ranging from those for optical flow estimation~\cite{Zhu2018} and visual odometry~\cite{Bryner2019} to those for object detection and tracking~\cite{Mitrokhin2018}. As an excellent overview is given in~\cite{Gallego2019}, we limit our survey to datasets that are potentially suitable for VPR. We further note that thus far there is only a single dataset captured by a color event camera, the Color Event Camera (CED) dataset~\cite{Scheerlinck2019CED}.

Many datasets are collected in a relatively small area, and their footage time is too short for VPR~\cite{Barranco2016, Bryner2019}. Other datasets are recorded over many hours in large areas but do not have associated location information~\cite{Cheng2019,Nitti2020}. The Multivehicle Stereo Event Camera Dataset \cite{Zhu2018} contains GPS information, but the sequence overlaps are too small to evaluate VPR performance.

To the best of our knowledge, the only dataset in the literature that contains multiple traverses of the same locations is the \ddddataset dataset~\cite{Binas2017}. It contains over 12 hours of recording in highway and city driving at different times of the day and in varying weather conditions. In Section~\ref{sec:experiments}, we evaluate two pairs of sequences where the same highway path was traversed. %

To foster future research on VPR using event cameras, we make our \datasetname dataset publicly available. %

\begin{figure*}[t]
  \vspace{.2cm}
  \centering
  \footnotesize
  \makebox[.03\linewidth]{}
  \makebox[.32\linewidth]{Single road}\hfil
  \makebox[.32\linewidth]{Dual carriageway}\hfil
  \makebox[.32\linewidth]{Shady areas}\hfil\\[0.1cm]
  \rotatebox[origin=c]{90}{RGB frames}
  \raisebox{-0.45\height}{
      \includegraphics[width=0.15\linewidth]{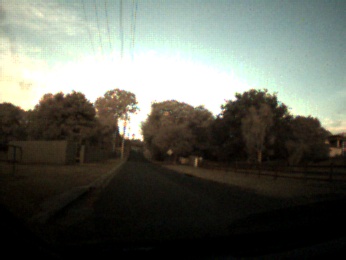}
      \includegraphics[width=0.15\linewidth]{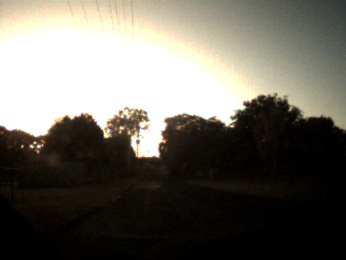}\hspace{0.01\linewidth}
      \includegraphics[width=0.15\linewidth]{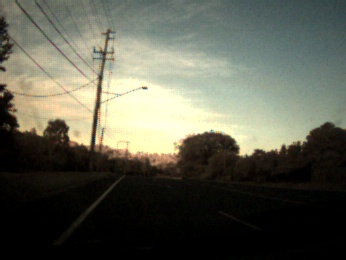}
      \includegraphics[width=0.15\linewidth]{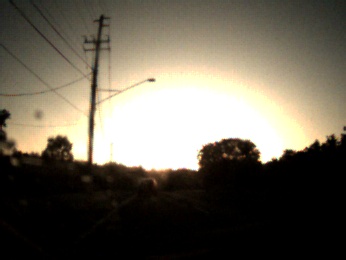}\hspace{0.01\linewidth}
      \includegraphics[width=0.15\linewidth]{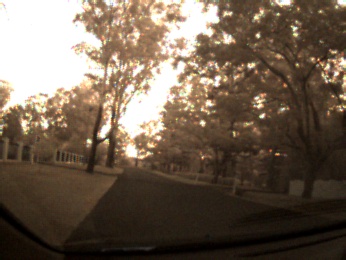}
      \includegraphics[width=0.15\linewidth]{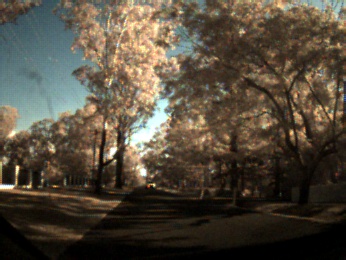}
  }\\[0.1cm]
  \rotatebox[origin=c]{90}{Reconstructed}
  \raisebox{-0.45\height}{
      \includegraphics[width=0.15\linewidth]{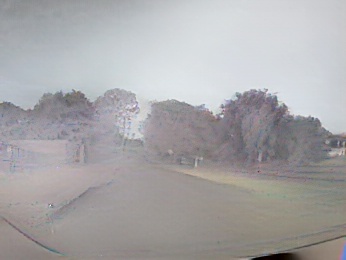}
      \includegraphics[width=0.15\linewidth]{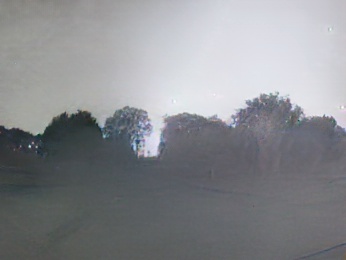}\hspace{0.01\linewidth}
      \includegraphics[width=0.15\linewidth]{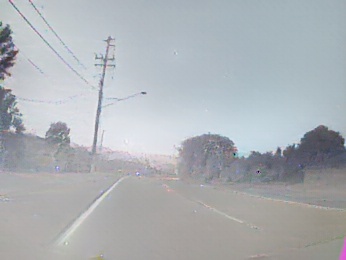}
      \includegraphics[width=0.15\linewidth]{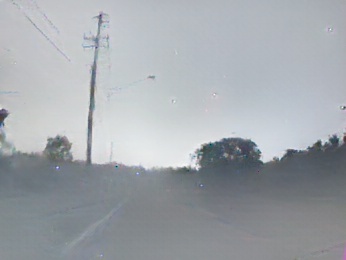}\hspace{0.01\linewidth}
      \includegraphics[width=0.15\linewidth]{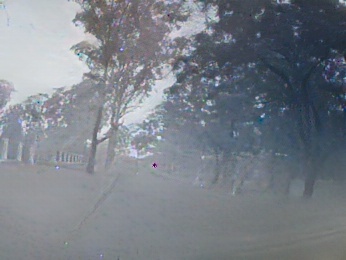}
      \includegraphics[width=0.15\linewidth]{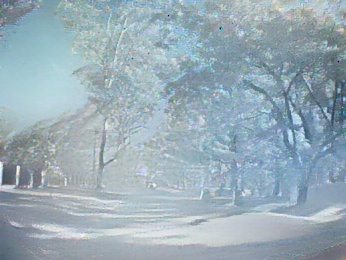}
  }
  \caption{Example of images contained in the \protect\datasetname dataset. The top row contains traditional DAVIS346 RGB camera frames, while the bottom row contains reconstructed images. We show pairs of images for two traverses that were captured at different times of the day in three environments: single road, dual carriageway, and shady areas. The Multimedia Material contains more images, including those captured by a consumer camera.}%
  \label{fig:example_images}%
  \vspace{-.2cm}
\end{figure*}

\section{Proposed Approach}
In this section, we first show how an event stream can be split into multiple windows that can be processed in parallel (Section~\ref{subsec:parallelprocessing}). We then present a processing pipeline for the purpose of visual place recognition (Section~\ref{subsec:vpr}). Finally, we show how multiple windows can be combined using an ensemble scheme (Section~\ref{subsec:ensemble}) and present a more computationally efficient, approximate ensemble version (Section~\ref{subsec:approximateensemble}).

\subsection{Parallel Window Processing}
\label{subsec:parallelprocessing}
The $i$-th event $\mathbf{e}_i = (\mathbf{u}_i,t_i,p_i)$ is described by the pixel coordinates $\mathbf{u}_i=(x_i,y_i)^T$ and time $t_i$ at which a brightness change of polarity $p_i\in\{-1,1\}$ was triggered. In a standard formulation, the events are then split into non-overlapping temporal windows $\bm{\epsilon}_k=\{\mathbf{e}_i\}$. The window size can either be chosen such that $N$ events are contained in each window, i.e.
\begin{equation}
    \lvert\bm{\epsilon}_k^N\rvert=N \text{ with } \bm{\epsilon}_k^N \cap \bm{\epsilon}_{k+1}^N = \varnothing,
\end{equation} or so that a fixed time span $\tau$ is considered, i.e.
\begin{equation}
    \bm{\epsilon}_k^\tau\coloneqq\{\mathbf{e}_i\mid t_i\in[t_k,t_k+\tau)\} \text{ with } t_{k+1} = t_k + \tau.    
\end{equation}

In this letter, we propose to consider a set of temporal windows $\mathbf{E}$ containing both $L$ fixed-size and $M$ fixed-time windows: 
\begin{equation}
    \mathbf{E}\coloneqq \bigcup\limits_{l=1}^{L} \bm{\epsilon}_{k_l}^{N_l} \cup \bigcup\limits_{m=1}^{M} \bm{\epsilon}_{k_m}^{\tau_m}.
\end{equation}

Each temporal window $\bm{\epsilon}_k$ is then processed individually as described in the next section, and treated as an ensemble member within an ensemble scheme as detailed in \mbox{Section~\ref{subsec:ensemble}}.

\subsection{Event-Based Visual Place Recognition}
\label{subsec:vpr}
\textbf{Image reconstruction:} For each $\bm{\epsilon}_k^n \in \mathbf{E}$, we first extract a corresponding image $\bm{\mathcal{I}}_k^n$. To this end, we employ the recent work by Rebecq et al.~\cite{Rebecq2019}, who propose E2VID, a recurrent, fully convolutional network similar to the UNet architecture~\cite{Ronneberger2015}. 

They train their architecture on simulated event sequences, which has the advantage that corresponding ground-truth images are available. The loss function is a combination of an image reconstruction loss and a temporal consistency loss. 

Note that our ensemble scheme does not rely on a particular image reconstruction method; indeed, in Section~\ref{subsec:ablation}, we will compare it to FireNet~\cite{Scheerlinck2020}, a faster reconstruction method using a much smaller network.

\textbf{Temporal alignment:} One problem that arises when using varying window lengths is that they do not align. To mitigate this problem, for the desired time $t^*$, we select the temporal windows that contain the event $\mathbf{e}_i$ with the closest timestamp $t_i$ to $t^*$: 
\begin{equation}
    \lvert t_i-t^*\rvert < \lvert t_j-t^*\rvert\ \quad \forall\ j \neq i.
\end{equation}

\textbf{Feature extraction and descriptor comparison:} Given reconstructed images $\bm{\mathcal{I}}_k^n$ sampled at time intervals $\Delta t$, we then extract features that can be used for VPR. NetVLAD~\cite{Arandjelovic2018} has been shown to be remarkably robust across multiple datasets and has since been extended to other tasks such as action classification~\cite{girdhar2017actionvlad}. The NetVLAD architecture consists of a convolutional neural network followed by a NetVLAD layer. Given an image $\bm{\mathcal{I}}_k^n$, the network outputs a 4096-dimensional feature descriptor $\mathbf{d}_k^n$.

Analogous to the image reconstruction method, our ensemble scheme does not rely on a particular feature extraction method. In Section~\ref{subsec:performanceourdataset}, we will demonstrate that our ensemble scheme can also be applied to match images via a direct pixel-based method: the sum of absolute differences (as used in SeqSLAM~\cite{Milford2012}).

\subsection{Ensemble Scheme}
\label{subsec:ensemble}
Before applying our ensemble scheme, we separately build distance matrices $\mathbf{D}_k$ for each temporal \mbox{window $k$} to compare each query image with all reference images. In the case of NetVLAD, the entry $(n_Q, n_R)$ of a distance matrix contains the cosine distance between the descriptor of the query image $\bm{\mathcal{I}}_k^{n_Q}$ and the reference candidate $\bm{\mathcal{I}}_k^{n_R}$:
\begin{equation}
    \mathbf{D}_k(n_Q, n_R) = 1 - \frac{\mathbf{d}_k^{n_Q} \cdot \mathbf{d}_k^{n_R}}{\lVert \mathbf{d}_k^{n_Q} \rVert_2\ \lVert \mathbf{d}_k^{n_R} \rVert_2}.
\end{equation}

Each $\mathbf{D}_k$ is then used as an ensemble member and the ensemble matrix $\mathbf{D}_E$ is defined by the mean values of all individual distance matrices (remember that $\mathbf{E}$ contains $L$ fixed-size and $M$ fixed-time windows):
\begin{equation}
    \mathbf{D}_E(n_Q, n_R) = \frac{1}{\lvert \mathbf{E} \rvert} \sum_{k=1}^{\lvert \mathbf{E} \rvert} \mathbf{D}_k(n_Q, n_R).
\end{equation}

Alternative combination schemes, including majority voting, are presented and evaluated in Section~\ref{subsec:ablation}. Note that the mean rule leads to the same results as the sum rule, as they are identical except for the normalization factor $1/\lvert \mathbf{E} \rvert$.

\subsection{Approximate Ensemble}
\label{subsec:approximateensemble}
One of the drawbacks of our proposed ensemble scheme is the computational burden that comes with the requirement of evaluating multiple window sizes. Here, we provide an alternative approach where the computational demands are relaxed by only using a single window size at query time. 

We propose to compare the query feature vector obtained from a single window size ($\widehat{\mathbf{d}}^{n_Q}$) with the set of reference feature vectors originating from different window sizes contained in the reference set ($\mathbf{d}_k^{n_R}$). We emphasize that the computational benefit lies in the fact that $\widehat{\mathbf{d}}^{n_Q}$ stems from a single temporal window (no dependence on $k$).

In this case, the additional computation is reduced to $\lvert \mathbf{E} \rvert$ feature vector comparisons, rather than having to extract~$\lvert \mathbf{E} \rvert$~feature vectors \emph{and} comparing $\lvert \mathbf{E} \rvert$ feature vectors.

Formally, the approximate ensemble $\widehat{\mathbf{D}}_E$ is obtained as:
\begin{align}
    \widehat{\mathbf{D}}_E(n_Q, n_R) &= \frac{1}{\lvert \mathbf{E} \rvert} \sum_{k=1}^{\lvert \mathbf{E} \rvert} \widehat{\mathbf{D}}_k(n_Q, n_R)\text{, where}\\
    \widehat{\mathbf{D}}_k(n_Q, n_R) &= 1 - \frac{\widehat{\mathbf{d}}^{n_Q} \cdot \mathbf{d}_k^{n_R}}{\lVert \widehat{\mathbf{d}}^{n_Q} \rVert_2\ \lVert \mathbf{d}_k^{n_R} \rVert_2}.
\end{align}

\section{\datasetname Collection}

In this section, we describe the collection of \datasetname (see Fig.~\ref{fig:example_images} for sample images). We first introduce our recording setup in Section~\ref{subsec:recordingsetup}, followed by an overview of the data that was recorded in Section~\ref{subsec:datastoverview}. Finally, we describe the ground truth annotation and post-processing steps in Sections~\ref{subsec:GTannotation} and~\ref{subsec:postprocessing}.

\vspace*{-0.2cm}
\subsection{Recording Setup}
\label{subsec:recordingsetup}
\begin{figure}[t]
  \centering
  \vspace{0.2cm}
  \begin{overpic}[width=\linewidth,tics=10]{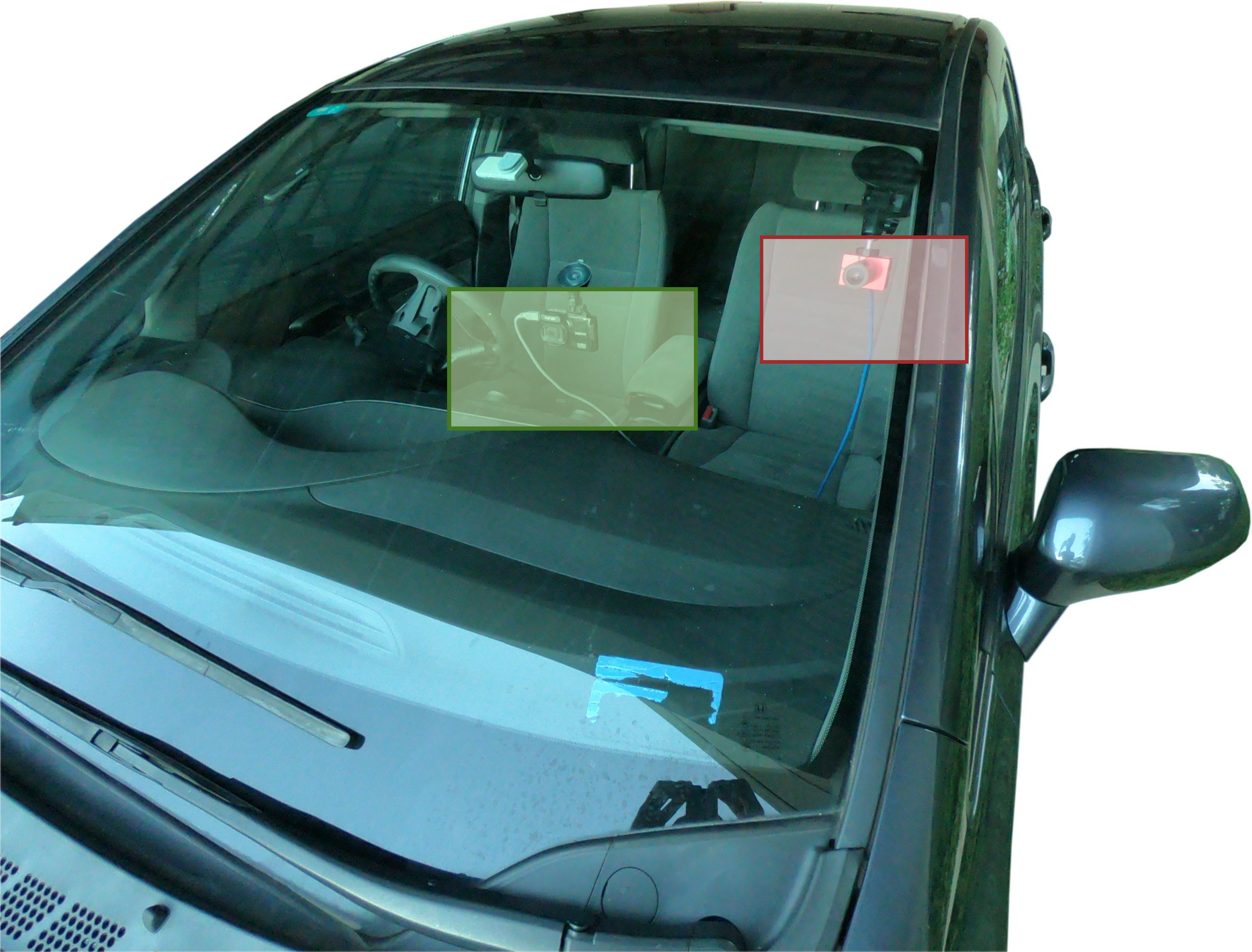}
    \put (39.5,45.5) {\scriptsize \textcolor{black}{{Consumer}}}
    \put (41.5,42.5) {\scriptsize \textcolor{black}{{Camera}}}
    \put (62.75,49) {\scriptsize \textcolor{black}{{DAVIS346}}}
  \end{overpic}
  \caption{Recording setup. The DAVIS346 was mounted on the windshield of a Honda Civic (red box). Another camera was used to collect RGB images and GPS information (green box).}%
  \label{fig:recording_setup}%
\end{figure}

We used a DAVIS346 color event camera~\cite{DAVIS346,Li2015} to record \datasetname. The DAVIS346 was mounted forward-facing on the inside of the windshield of a Honda Civic, as shown in Fig.~\ref{fig:recording_setup}. The DAVIS346 allows recording of events and aligned RGB frames with 346x260 pixels resolution. We used the same bias settings as in the \ddddataset dataset~\cite{Binas2017}\footnote{We note that these bias settings led to significantly better results than the standard bias settings or those used to record the Color Event Dataset~\cite{Scheerlinck2019CED} in our scenario.} and used the RPG ROS DVS package~\cite{Mueggler2014} as the recording tool. The DAVIS346 provides color information for each event using a color filter array (neighboring pixels represent different colors)~\cite{Li2015}. We also provide the readings of the built-in Inertial Measurement Unit (IMU).

While not used in this letter, our \datasetname dataset also contains videos captured by a consumer camera in 1080p resolution and GPS information (see Multimedia Material for example images). The frame-based camera is mounted below the reverse mirror (see Fig.~\ref{fig:recording_setup}).

\vspace*{-0.2cm}
\subsection{Dataset Overview}
\label{subsec:datastoverview}
\begin{figure}[t]
  \centering
  \vspace{0.2cm}
  \includegraphics[width=0.72\linewidth]{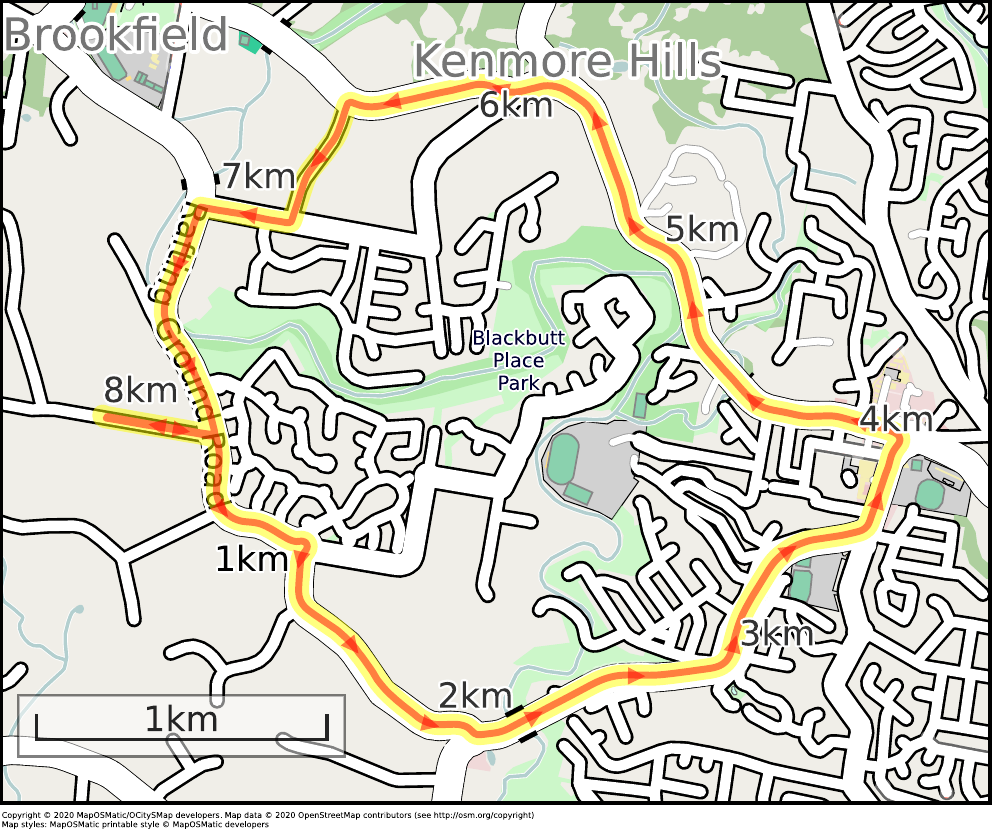}
  \caption{Route in the \datasetname dataset. The route is approx.~8km long and was traversed \numtraverses times at different times of the day.}%
  \label{fig:dataset_map}%
  \vspace*{-0.2cm}
\end{figure}

The dataset was captured in the Brookfield and Kenmore Hills outer suburbs of Brisbane (see Fig.~\ref{fig:dataset_map}). The route is approx.~8km long and contains a variety of different roads, ranging from single-lane roads without much build-up over dual carriageways to built-up areas. Some areas contain many trees that cast shadows on the street and lead to challenging lighting conditions. The dataset includes \numtraverses traverses recorded at different times of the day and under varying weather conditions. %

\vspace*{-0.05cm}
\subsection{Ground Truth Annotation}
\label{subsec:GTannotation}
We manually annotated ground truth matches based on distinct landmarks (traffic signs, street lights, traffic intersections, etc.) on average every 15-20s (approx.~every 200-300m). We removed stationary periods and then used linear interpolation between the ground truth matches to find one match per second ($\Delta t=1s$).

\vspace*{-0.05cm}
\subsection{Post-processing}
\label{subsec:postprocessing}
We first remove hot pixels using the `dvs\_tools'\footnote{\url{https://github.com/cedric-scheerlinck/dvs_tools}}. We then use a simple filter to remove bursts in the event stream (events that are erroneously triggered for the majority of pixels within a very short time), which is a known issue with the DAVIS346 camera when sunlight hits the internal bias generator. This postprocessing is consistent with other research in the field and is a symptom of the relative immaturity of this technology compared to, for example, RGB consumer cameras. We include this burst filter in our code repository.

\section{Experimental Results}
\label{sec:experiments}
This section presents the experimental results. We first describe the experimental protocol and performance metric in Section~\ref{subsec:implementation}. 
The performance on our \datasetname dataset is evaluated in Section~\ref{subsec:performanceourdataset}, and is followed by results on the \ddddataset dataset in Section~\ref{subsec:performance-ddd-17}. We then present ablation studies in Section~\ref{subsec:ablation} with regards to 1) the image reconstruction method used, 2) the strategies that can be used to combine the ensemble members, and 3) a comparison to a model-based ensemble.

\begin{figure*}[t]
  \vspace{0.2cm}
  \makebox[\linewidth]{\resizebox{.6\linewidth}{!}{
    \includegraphics{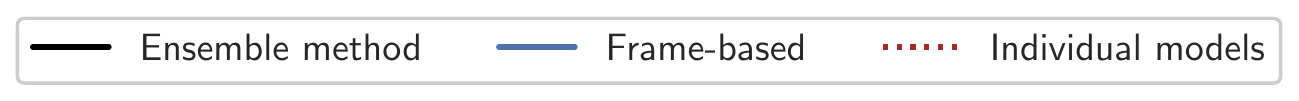}
  }}\\
  \resizebox{!}{.165\linewidth}{
    \begin{adjustbox}{clip,trim=0.25cm 0.8cm 0.25cm 0.25cm}
      \includegraphics{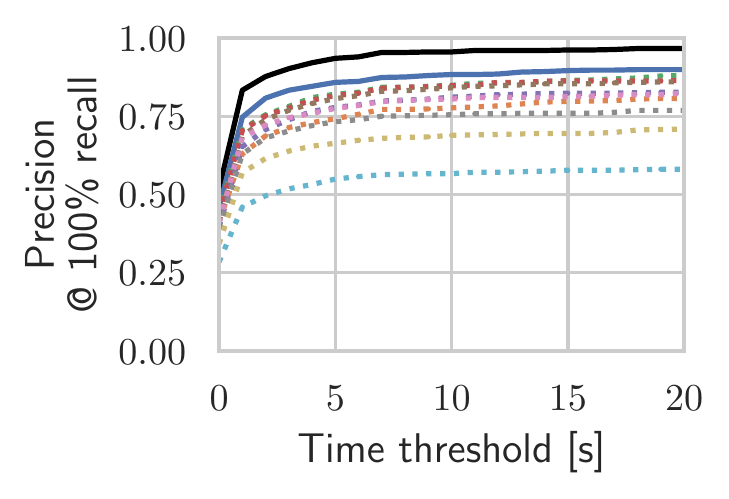}
    \end{adjustbox}
  }
  \hspace{0.1cm}
  \hfil
  \resizebox{!}{.165\linewidth}{
    \begin{adjustbox}{clip,trim=2.1cm 0.8cm 0.25cm 0.25cm}
      \includegraphics{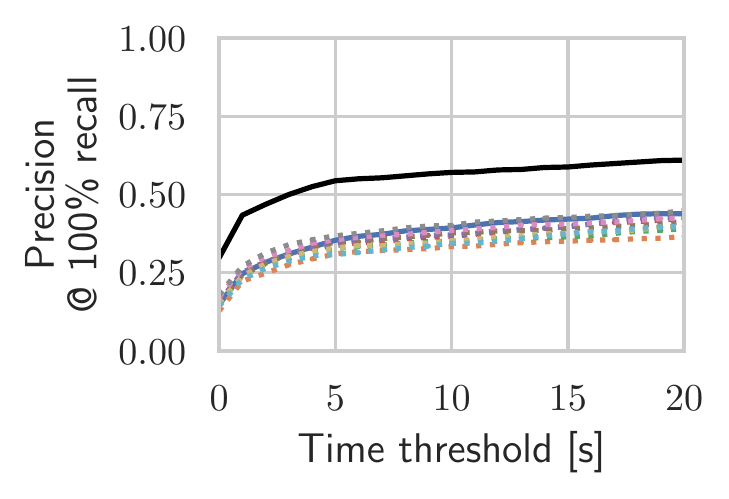}
    \end{adjustbox}
  }
  \hspace{0.1cm}
  \hfil
  \resizebox{!}{.165\linewidth}{
    \begin{adjustbox}{clip,trim=2.1cm 0.8cm 0.25cm 0.25cm}
      \includegraphics{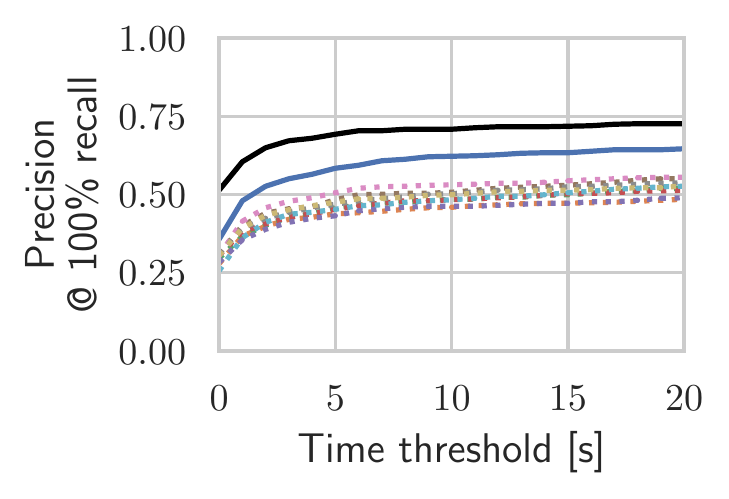}
    \end{adjustbox}
  }
  \hspace{0.1cm}
  \hfil
  \resizebox{!}{.165\linewidth}{
    \begin{adjustbox}{clip,trim=2.1cm 0.8cm 0.25cm 0.25cm}
      \includegraphics{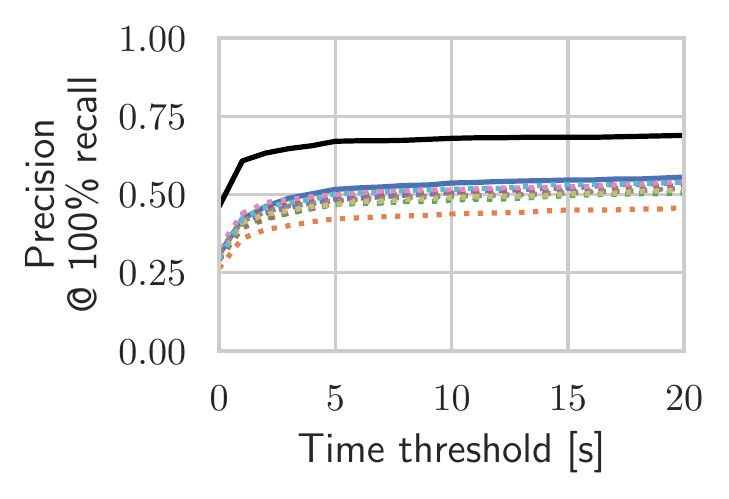}
    \end{adjustbox}
  }
  \\
  \resizebox{!}{.186\linewidth}{
    \begin{adjustbox}{clip,trim=0.25cm 0.25cm 0.25cm 0.25cm}
      \includegraphics{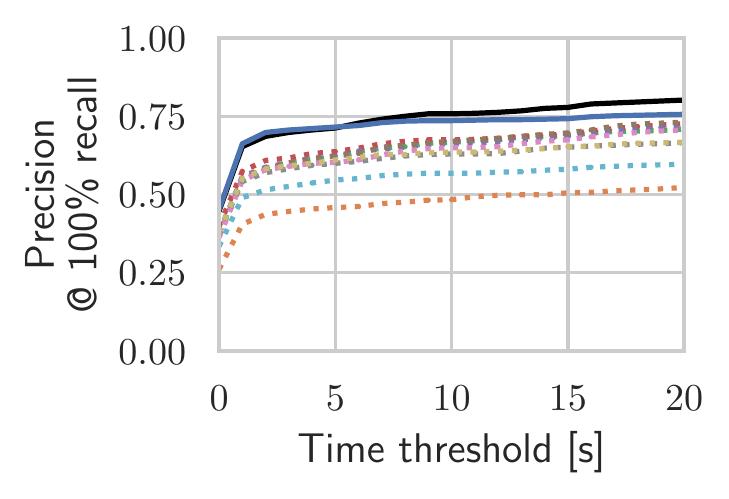}
    \end{adjustbox}
  }
  \hfil
  \resizebox{!}{.186\linewidth}{
    \begin{adjustbox}{clip,trim=2.1cm 0.25cm 0.25cm 0.25cm}
      \includegraphics{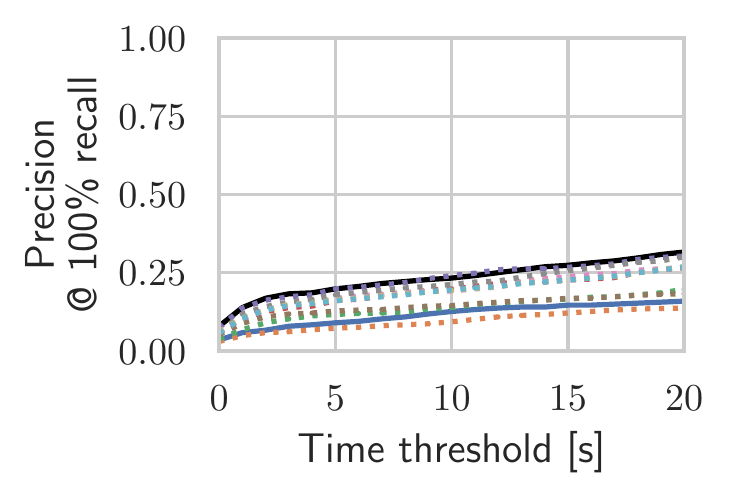}
    \end{adjustbox}
  }
  \hfil
  \resizebox{!}{.186\linewidth}{
    \begin{adjustbox}{clip,trim=2.1cm 0.25cm 0.25cm 0.25cm}
      \includegraphics{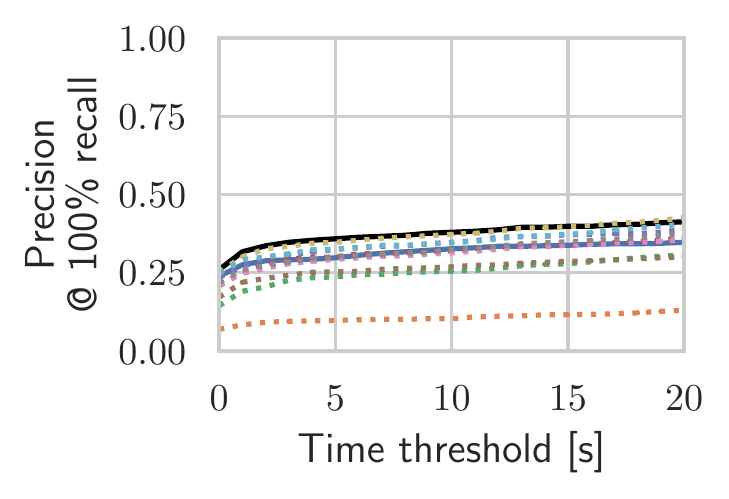}
    \end{adjustbox}
  }
  \hfil
  \resizebox{!}{.186\linewidth}{
    \begin{adjustbox}{clip,trim=2.1cm 0.25cm 0.25cm 0.25cm}
      \includegraphics{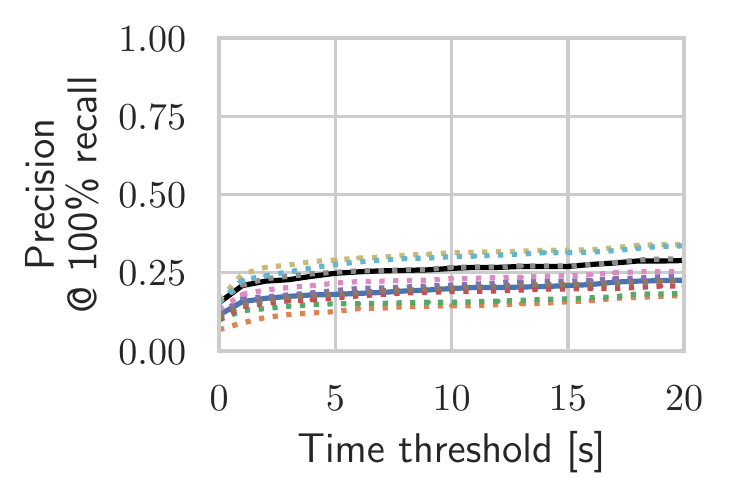}
    \end{adjustbox}
  }
  \hfil
  \small
  \makebox[.08\linewidth]{}\makebox[.21\linewidth]{Sunset 1 vs sunset 2}\makebox[.26\linewidth]{Sunset 1 vs day-time}\makebox[.21\linewidth]{Sunset 1 vs morning}\makebox[.07\linewidth]{}\makebox[.132\linewidth][r]{Sunset 1 vs sunrise}
  \vspace*{-0.1cm}
  \caption{Performance comparison on \datasetname for matching images under varying environmental conditions using NetVLAD as feature extractor (top row) or Sum of Absolute Differences (bottom row). In all but one case the ensemble (black) outperforms all individual ensemble members (dotted lines), independent of the localization threshold. The ensemble also outperforms the frame-based method (blue). A summary plot can be found in Fig.~\ref{fig:barplot}.}%
  \label{fig:recall_varying_threshold}%
  \vspace*{-0.2cm}
\end{figure*}

\subsection{Experimental Protocol}
\label{subsec:implementation}
We used Python to implement \methodname. We found no significant performance differences for the ensemble scheme depending on the particular values of $N_l$ and $\tau_m$. For reference, we used %
$\widetilde{N} = \{ 0.1,\allowbreak 0.3,\allowbreak 0.6,\allowbreak 0.8\}$, where $\widetilde{N}$ is the normalized value for a resolution of $W\cdot H$\footnote{For the DAVIS346, $W=346$ and $H=260$ pixels.} such that $N=\widetilde{N}\cdot W\cdot H$, and $\tau= \{ 44,\allowbreak 66,\allowbreak 88,\allowbreak 120,\allowbreak 140 \}$ (ms).
We use precision at 100\% recall to evaluate the performance of our method. A match for a query is deemed as a true positive when its cosine distance is less than a similarity threshold, given a localization threshold of 5s\footnote{At an average speed of 50km/h, the car travels around 70m within 5s.}. The Multimedia Material includes further details, including precision-recall curves. %

To ensure a fair comparison, all window-specific methods and the ensemble scheme were treated equally in terms of the ground truth matches that were used for comparisons, performance evaluation, parameter settings, \textit{etc.}

\subsection{Comparison on the \datasetname dataset}
\label{subsec:performanceourdataset}
\textbf{Ensemble scheme performance:} Fig.~\ref{fig:recall_varying_threshold} shows the performance of \methodname on our \datasetname dataset. The ensemble scheme performs significantly better than all window-specific methods (paired $t$-tests, $p<0.01$ for all methods; using NetVLAD features). The average performance increase is $\performanceimprovement\pm\performanceimprovementstd\%$, and the improvement over the best performing individual ensemble method is $\performanceimprovementoverbest\pm\performanceimprovementoverbeststd\%$. %

\textbf{Feature extractor comparison:} Using NetVLAD as the feature extractor significantly outperforms the Sum of Absolute Differences. However, we note that for both NetVLAD features and the Sum of Absolute Differences, the ensemble scheme generally outperforms the individual ensemble members (from here on referred to as ``individual models''). For the remainder of this section, we therefore discuss results obtained using NetVLAD features.

\begin{figure}[t]
  \centering
  \includegraphics[width=.9\linewidth]{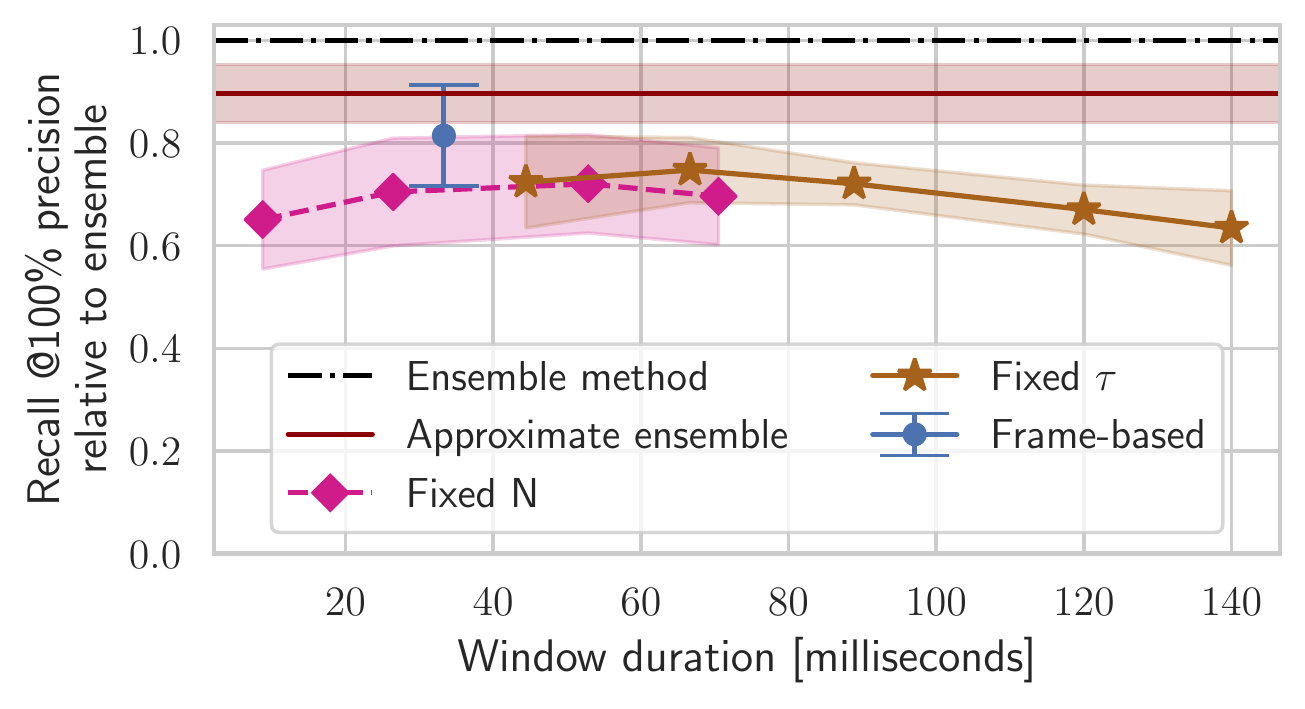}
  \caption{Performance on \datasetname with varying window duration. The performance of the individual models is normalized with respect to the ensemble performance and averaged over all sequences (standard deviation is shown as shaded area). Models relying on a fixed number of events in each window (pink; the median time each window spans was used on the $x$-axis) perform similar to those with a fixed time (brown). While there is a performance peak around 60ms, the standard deviation is quite large, so that for some sequences smaller windows perform better, while for others larger windows perform better. All results were obtained using NetVLAD features.\vspace*{-0.2cm}}%
  \label{fig:recall_varying_window}%
\end{figure}

\textbf{Performance with varying window duration:} Fig.~\ref{fig:recall_varying_window} shows the performance with varying window duration. One can observe that the performances of models based on a fixed number of events in each window and those based on a fixed time span follow a similar trend. In fact, except for $\widehat{N}=0.1$ (which performs worse), there is no significant performance difference between the individual models ($t$-tests, $p>0.1$).

\begin{figure}[t]
  \centering
  \includegraphics[width=.985\linewidth]{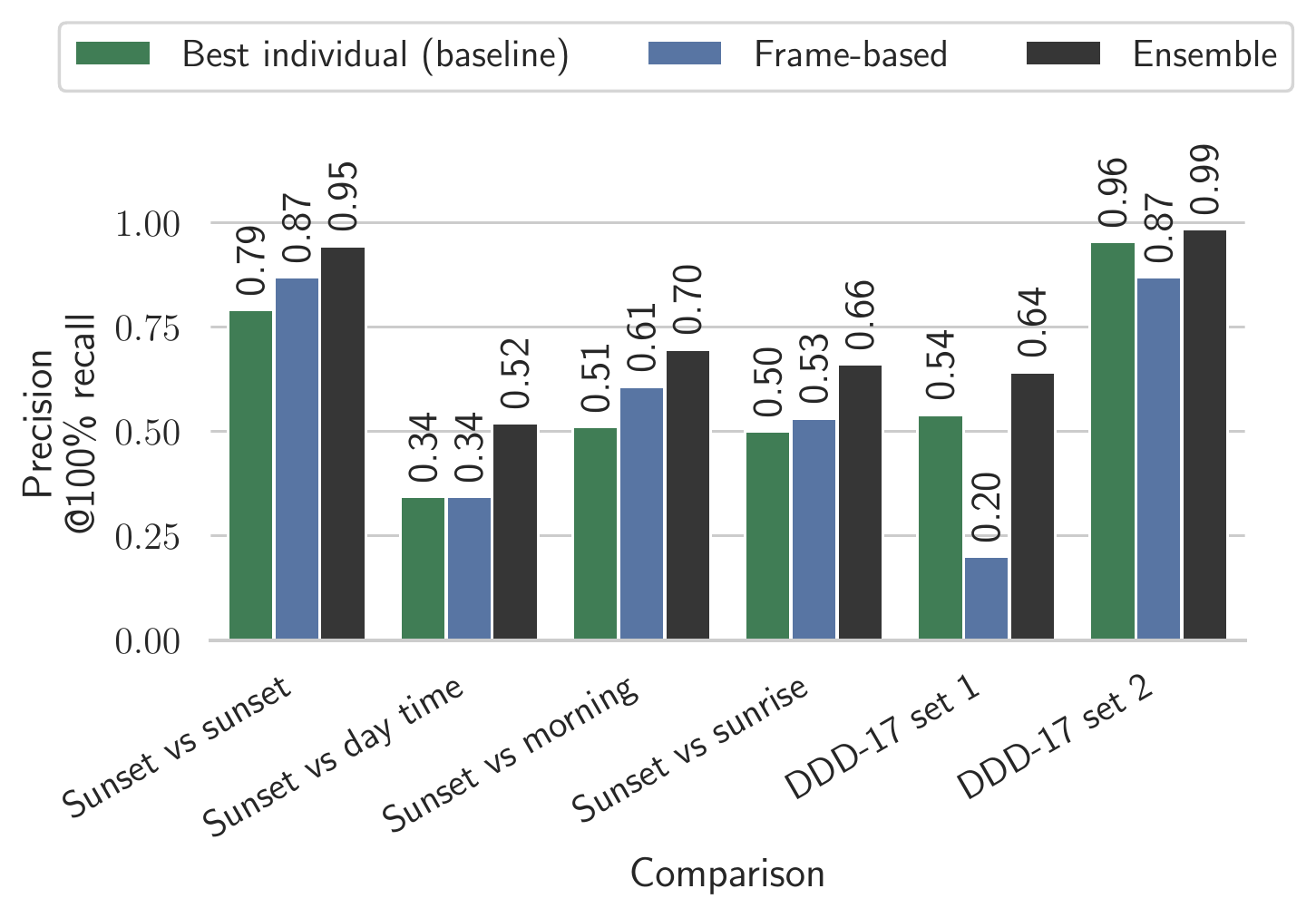}
  \vspace*{-.3cm}
  \caption{Absolute precision @100\% recall performance on the \datasetname and \ddddataset datasets for the best individual model (baseline), NetVLAD applied on the DAVIS346 RGB frames, and the ensemble scheme.}%
  \vspace*{-0.2cm}
  \label{fig:barplot}%
\end{figure}

\textbf{Comparison to features obtained from RGB frames:} Figs.~\ref{fig:recall_varying_window} and~\ref{fig:barplot} show that the ensemble method outperforms the features obtained by the DAVIS346 RGB frames (remember that the DAVIS346 can be used to record events \emph{and} RGB frames at the same time), which is particularly interesting considering that the features obtained using DAVIS346 RGB frames outperform all individual models. However, the DAVIS346 RGB frames are of relatively poor quality compared to consumer and high-end frame-based cameras. As our dataset also contains images recorded by a consumer camera, a comparison to better quality RGB frames can easily be performed in future works.

\textbf{Approximate ensemble scheme:} We now evaluate the approximate ensemble scheme where only a single temporal window is considered for the query set (see Section~\ref{subsec:approximateensemble}). We chose $\widehat{N}=0.5$ as the temporal window size; however, the specific choice of the temporal window does not significantly impact the results.

The approximate ensemble scheme performs $\performanceworseapproximate\pm\performanceworseapproximatestd\%$ worse than the ensemble scheme. However, that is still significantly better than the best individual model ($\performanceimprovementapproximatebest\pm\performanceimprovementapproximatebeststd\%$ performance increase).

\textbf{Day-vs-night scenario:} %
In the day-vs-night scenario, none of the individual models performs better than random guessing. Consequently, the ensemble averages random guesses, which itself results in a random guess. NetVLAD applied to the DAVIS346 RGB frames results in random guesses, too, indicating that the failure is not due to the event-based representation but due to the extreme appearance change.

The day-vs-night scenario thus shows that (at least one of) the individual models need to perform at above random chance for the ensemble scheme to work. We plan to tackle this challenging scenario using feature descriptors that were trained on event data using an end-to-end architecture, as outlined in Section~\ref{sec:conclusions}.

\subsection{Comparison on the \ddddataset dataset}
\label{subsec:performance-ddd-17}

\textbf{Traverses used in the \ddddataset dataset:} To show that our method generalizes to other datasets, we now present results obtained on the \mbox{\ddddataset} dataset~\cite{Binas2017}. Using GPS information, we found two sets of traverses suitable for VPR within \ddddataset: the first set captured on a freeway in the late evening in rainy conditions (query images) and at daytime in the second traverse (reference images), and the other set captured in the evening in the first traverse (query images) and at daytime in the second traverse (reference images)\footnote{Specifically, `rec1487350455' and `rec1487417411' in the first set, and `rec1487779465' and `rec1487782014' in the second set. Several other routes are traversed in opposite directions, which is outside the scope of this letter (but see~\cite{Garg2019ICRA} for a possible way of tackling VPR for opposing viewpoints).}. We annotated the matching traverses in the same way as for the \datasetname dataset (see Section~\ref{subsec:GTannotation}).

\begin{figure}[t]
  \centering
  {
   \vspace{0.2cm}
   \footnotesize
   \makebox[.333\linewidth]{Query}\hfil\makebox[.333\linewidth]{Reference GT}\hfil\makebox[.333\linewidth]{Reference match}\\
  }
  \includegraphics[width=0.95\linewidth]{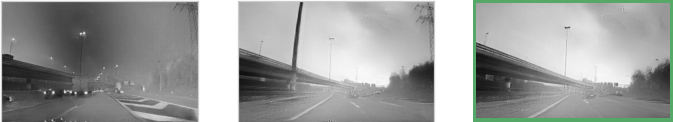}\\[0.05cm]
  \includegraphics[width=0.95\linewidth]{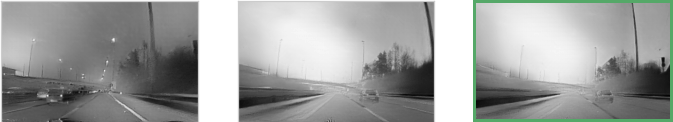}\\[0.2cm]
  \includegraphics[width=0.95\linewidth]{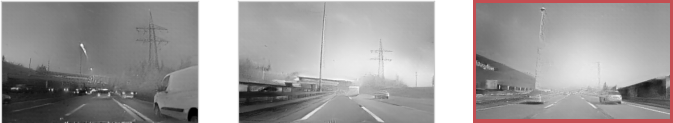}\\[0.05cm]
  \hfil\includegraphics[width=0.95\linewidth]{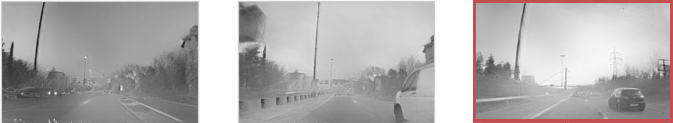}\hfil
  \vspace*{-0.18cm}
  \caption{Example matches of the ensemble and ground-truth (GT) matches on the \ddddataset dataset. Top two rows: success cases where the majority of individual methods failed. Bottom two rows: failure cases.}%
  \label{fig:query_gt_match}%
\end{figure}

\textbf{Performance on \ddddataset:} The general performance trend follows that presented for \datasetname. Specifically, the ensemble outperforms the individual models on average by $\performanceimprovementdddsetone\pm\performanceimprovementdddsetonestd\%$ in the first set and $\performanceimprovementdddsettwo\pm\performanceimprovementdddsettwostd\%$ in the second set. The ensemble also performs better than the best individual model ($\performanceimprovementoverbestdddsetone\%$ and $\performanceimprovementoverbestdddsettwo\%$ performance increase in the first and second set, respectively). We provide qualitative results obtained on the DDD-17 dataset in Fig.~\ref{fig:query_gt_match}.

Note that the relative improvements are not as large as in \datasetname, as the mean absolute performance is much higher ($\performancemeanddd\%$ precision @100\% recall in the \ddddataset dataset compared to $\performancemean\%$ in our dataset for the best performing individual model; see Fig.~\ref{fig:barplot}). As our ensemble ``filters out the noise'', the relative performance increases more in those cases where individual performances are lower.

Fig.~\ref{fig:barplot} also shows that performance obtained using RGB frames on DDD-17 is lower than that of the individual models, which is contrary to the performance on \datasetname.

\subsection{Ablation studies}
\label{subsec:ablation}

\textbf{Image reconstruction method:} We now briefly show that our ensemble scheme is agnostic to a particular image reconstruction method. While the aforementioned results were obtained using E2VID~\cite{Rebecq2019}, we also evaluated the performance when using FireNet~\cite{Scheerlinck2020} as the reconstruction method, a much smaller and faster architecture compared to E2VID. We found that there is neither a statistically significant difference in the absolute performance of individual models ($p>0.1$) nor a significant performance difference of the ensemble models ($p>0.1$).

\textbf{Comparison to other ensemble combination strategies:} Besides the mean rule, the majority voting scheme is one of the most commonly used strategies to combine ensemble members~\cite{Polikar2012}. In this scheme, the distance matrices $\mathbf{D}_k$ are combined within $\mathbf{D}_{MV}(n_Q, n_R)$, which is a zero matrix with ones at the modal value of $\{ \argmin{{n_R}} \mathbf{D}_k(n_Q,\ n_R), \forall n_Q \}$. While the majority voting scheme outperforms the individual models, it performs $\performanceworsemaxvote\pm\performanceworsemaxvotestd\%$ worse than the mean rule used in the proposed ensemble scheme. 

We argue that majority voting does not perform favorably as it only considers the highest-ranking match and thus might disregard other strong matches. More generally, majority voting does not consider the ``goodness'' (magnitude) of a match. The mean rule also finds ``near misses'' in the ensemble members, which is not the case for majority voting.

We also investigated the product, median, min, max, trimmed mean and weighted average rules~\cite{Polikar2012}, and similarly found that they do not perform as well as the mean rule (see Multimedia Material). %

\textbf{Comparison to a model-based ensemble:} We now compare to a more typical ``model-based'' ensemble, where two ensemble members are obtained, one by using E2VID~\cite{Rebecq2019} as the image reconstruction method and the other one by using FireNet~\cite{Scheerlinck2020}. Contrary to our ensemble that uses temporal windows of varying sizes, the model-based ensemble is based on a single window size ($\tau=55ms$, however, as noted earlier, the choice of the window size does not significantly impact performance). 

While the model-based ensemble also performs significantly better than each of the ensemble members ($\performanceimprovementmodel\pm\performanceimprovementmodelstd\%$ performance improvement), the performance improvement is not as significant as when using temporal ensembles. It will be interesting to build model-based ensembles with more members as soon as further image reconstruction methods become available, e.g.~\cite{stoffregen2020train}.

\section{Conclusions}
\label{sec:conclusions}
In this letter, we introduced an ensemble scheme to combine temporal event windows of different lengths. We have demonstrated that the ensemble scheme proves particularly effective in the visual place recognition task. To allow for large-scale evaluations of our method, we recorded the \datasetname dataset, which we make available to the research community. We note that our dataset contains the longest footage recorded using a color event camera thus far (\datasetlengthminutes\ minutes). We continue to record more traverses of the same route and will add a second route in the future.

Our future works are made up of four pillars: 
1)~We want to explore the relationship between the car velocity and the performance of individual ensemble members. We expect that shorter time windows are required for large velocities and vice versa. Therefore, the car velocity could be used as a prior to inform the temporal window sizes of ensemble members and to reduce computational requirements by using fewer ensemble members.
2)~We seek to demonstrate that our proposed ensemble scheme generalizes to other tasks such as feature tracking, visual-inertial odometry, and steering prediction in driving. 
3)~Another tantalizing possibility would be to train an end-to-end architecture, following the key work by Gehrig et al.~\cite{Gehrig2019}. Training data for this architecture will be obtained using recent research that allows creation of virtual events from normal RGB footage~\cite{Gehrig2020}.
4)~A trainable architecture will then enable the use of more complex ensemble learning methods such as a mixture of experts, where individual ensemble members learn to become experts of some portion of the feature space~\cite{jacobs1991adaptive}. Such an approach is particularly interesting in VPR, where one can imagine scene-specific experts~\cite{Choi_2018_CVPR}.

{
\hbadness 10000\relax
\bibliographystyle{IEEEtran}
\bibliography{references,references_ieeectrl}
}

\end{document}


\bstctlcite{MyBSTcontrol}

\maketitle

\section*{Additional Results}
\subsection{Comparison to other ensemble combination strategies}
\begin{figure}
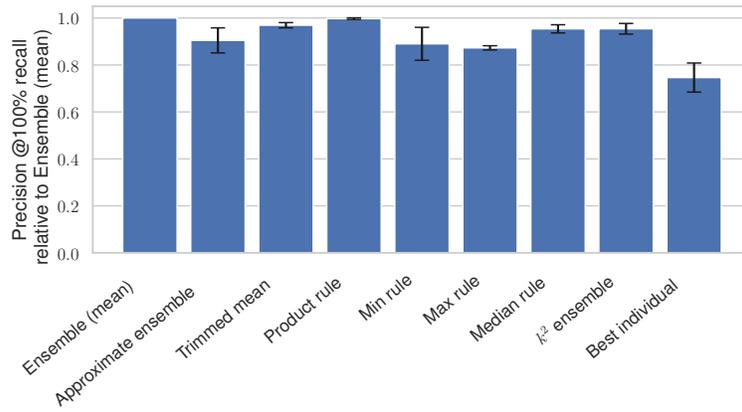

  \centering
  \resizebox{!}{.32\linewidth}{
     \inputpgf{figs}{performance_across_windows_ensemble_types_netvlad.pgf}
  }
  \caption{Comparison of alternative ensemble strategies. All ensemble strategies perform better than the best individual model. The proposed ensemble scheme (mean rule) performs similarly when compared to the product rule, and better than the other ensemble strategies.}%
  \label{fig:comparison_ensemble_strategies}%
\end{figure}

Suppl.~Fig.~\ref{fig:comparison_ensemble_strategies} provides performance comparisons for the product, median, min, max, and trimmed mean rules as described in Section V-D of the main paper. All ensemble strategies perform better than the best individual models, and the mean and product rules perform slightly better than the other ensemble strategies.

Furthermore, we considered weighted averages. For this, we introduced weights $\alpha_k \in \{0.5, 0.75, 1.0, 1.25, 1.5\}$ such that:
\begin{equation}
    \mathbf{D}_E(n_Q, n_R) = \frac{1}{\lvert \mathbf{E} \rvert} \sum_{k=1}^{\lvert \mathbf{E} \rvert} \alpha_k \mathbf{D}_k(n_Q, n_R)
\end{equation}
and considered all possible combinations of $\alpha_k$. We found that weighted averages do not result in better performances as each ensemble member's contribution heavily varies across different traverses. In future works, we would like to explore whether training $\alpha_k$ on a training set leads to favorable performances. 

\subsection{Combination of two ensemble methods}
Section III-C of the letter introduces our ensemble scheme. The main idea is to compare query features obtained using a specific window size with the reference features obtained using the same window size, and do this for $k$ temporal window sizes. Here we explore whether it is beneficial to also compare features that were obtained from a different window size, such that the ensemble contains $k^2$ members (one member for each possible combination of temporal windows)\footnote{We would like to thank the anonymous reviewer for encouraging us to include these results.}.

We find that this combined ensemble strategy performs $5.1\pm2.7\%$ worse than our proposed ensemble scheme. This indicates that the features of ``shorter'' time windows are not compatible with those of ``longer'' time windows. However, a more thorough investigation of this hypothesis is required in future works. The performance in comparison to other ensemble strategies is shown in Suppl.~Fig.~\ref{fig:comparison_ensemble_strategies}.

\subsection{Additional dataset images}
\begin{figure*}[t]
  \vspace{.2cm}
  \centering
  \footnotesize
  \makebox[.07\linewidth]{}
  \makebox[.38\linewidth]{Pedestrian crossing}\hfil
  \makebox[.53\linewidth]{Traffic light}\\[0.1cm]
  \rotatebox[origin=c]{90}{RGB frames}
  \raisebox{-0.45\height}{
      \includegraphics[width=0.22\linewidth]{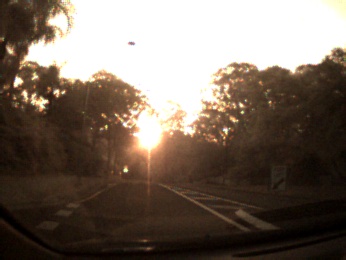}
      \includegraphics[width=0.22\linewidth]{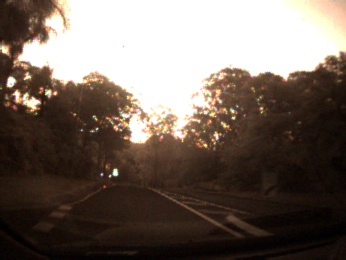}\hspace{0.01\linewidth}
      \includegraphics[width=0.22\linewidth]{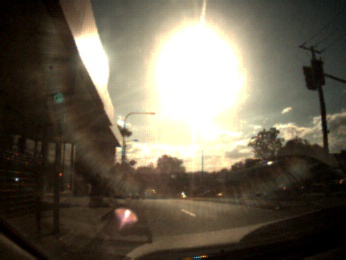}
      \includegraphics[width=0.22\linewidth]{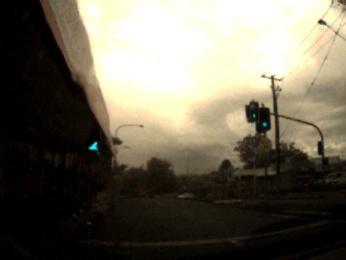}
  }\\[0.1cm]
  \rotatebox[origin=c]{90}{Reconstructed}
  \raisebox{-0.45\height}{
      \includegraphics[width=0.22\linewidth]{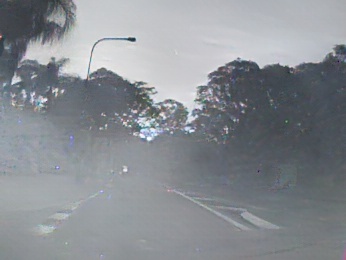}
      \includegraphics[width=0.22\linewidth]{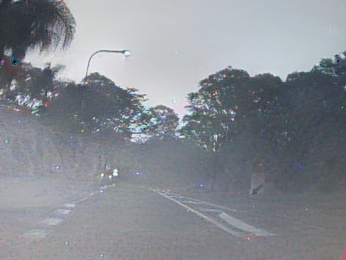}\hspace{0.01\linewidth}
      \includegraphics[width=0.22\linewidth]{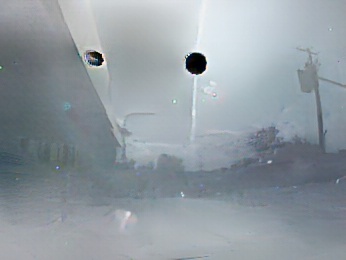}
      \includegraphics[width=0.22\linewidth]{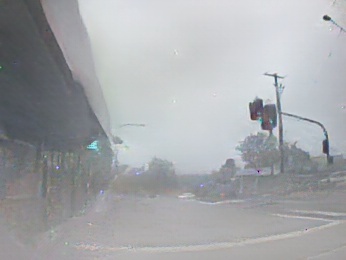}
  }\\[0.1cm]
  \rotatebox[origin=c]{90}{Consumer cam.}
  \raisebox{-0.45\height}{
      \includegraphics[width=0.22\linewidth]{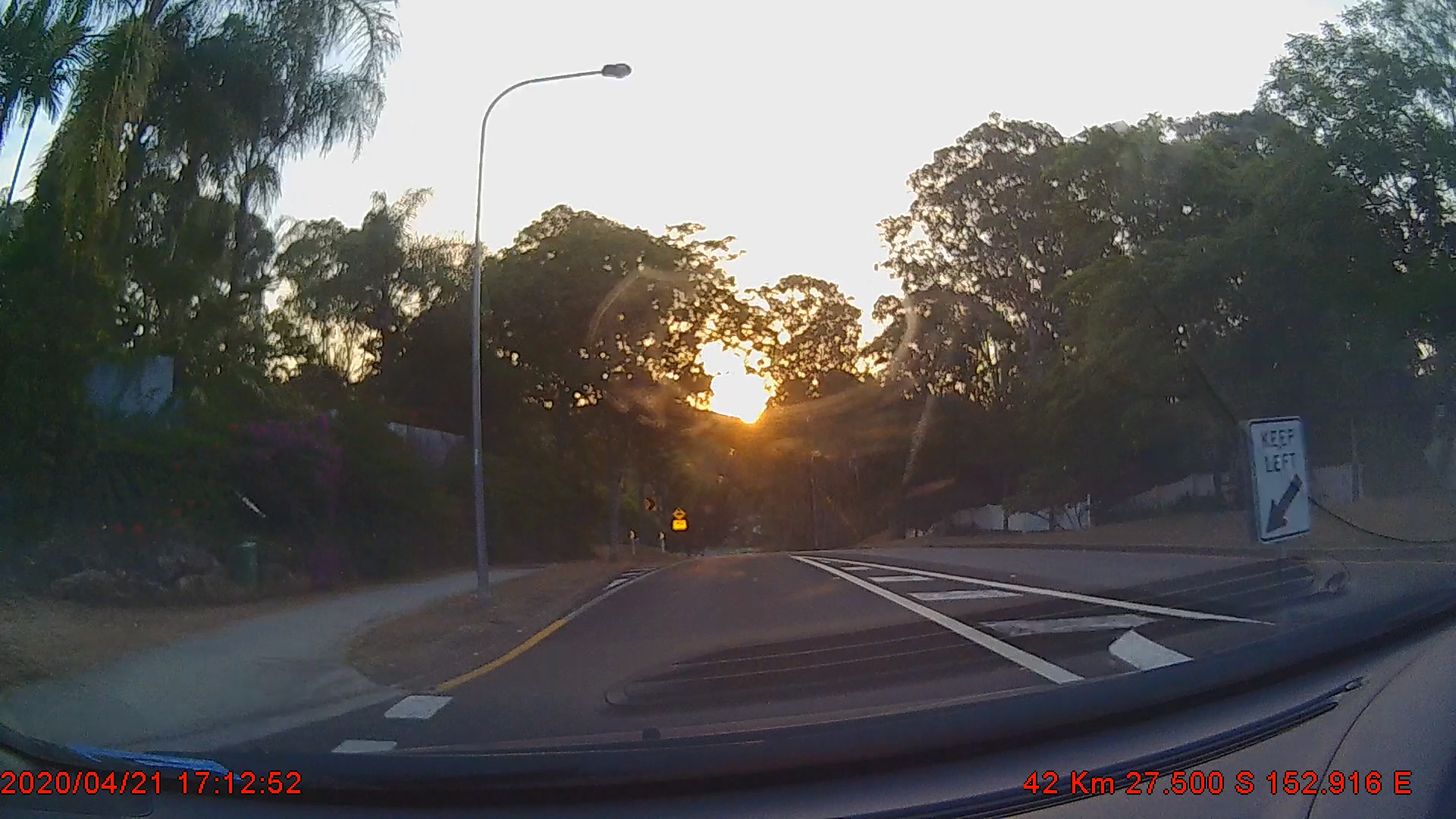}
      \includegraphics[width=0.22\linewidth]{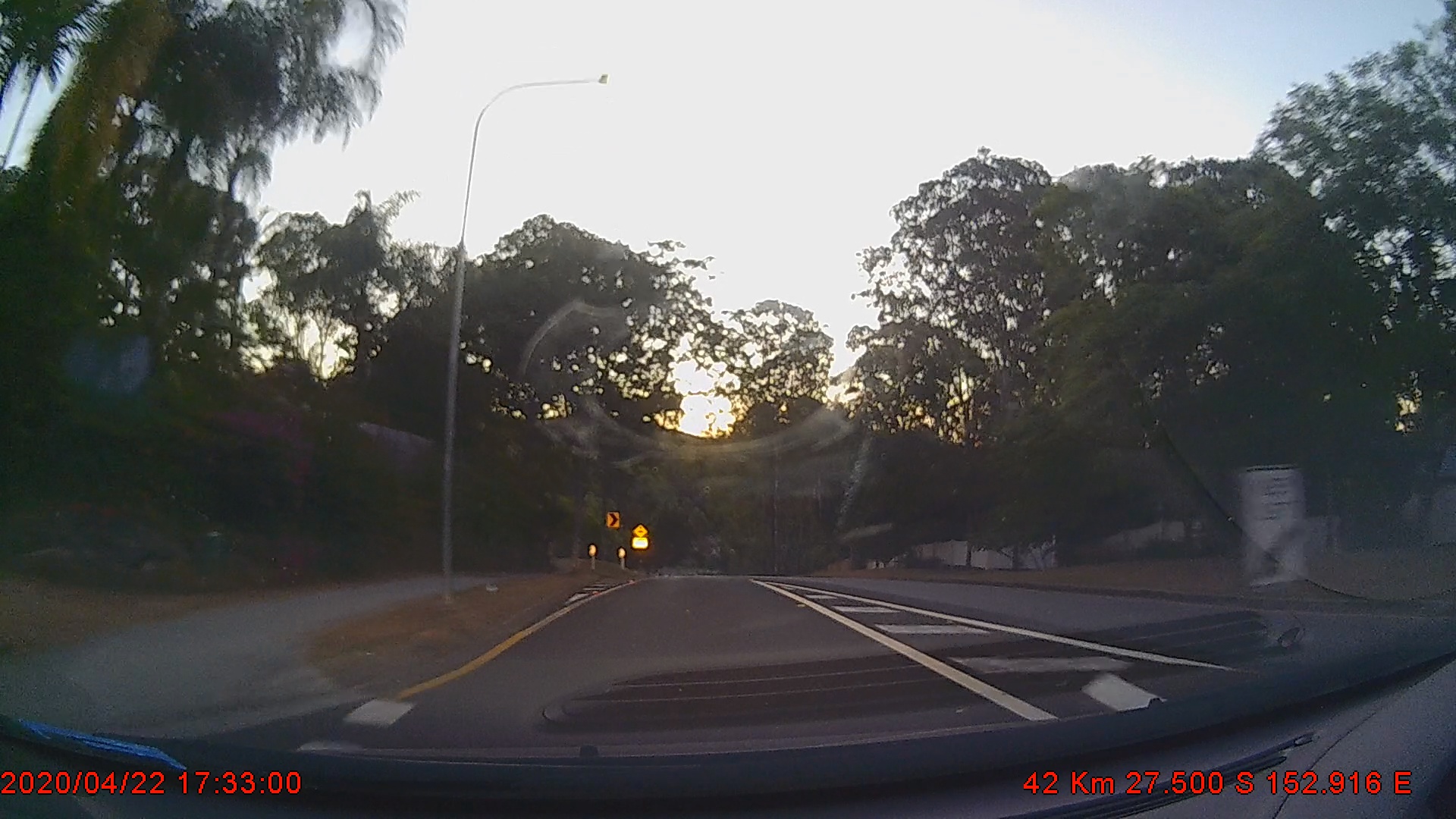}\hspace{0.01\linewidth}
      \includegraphics[width=0.22\linewidth]{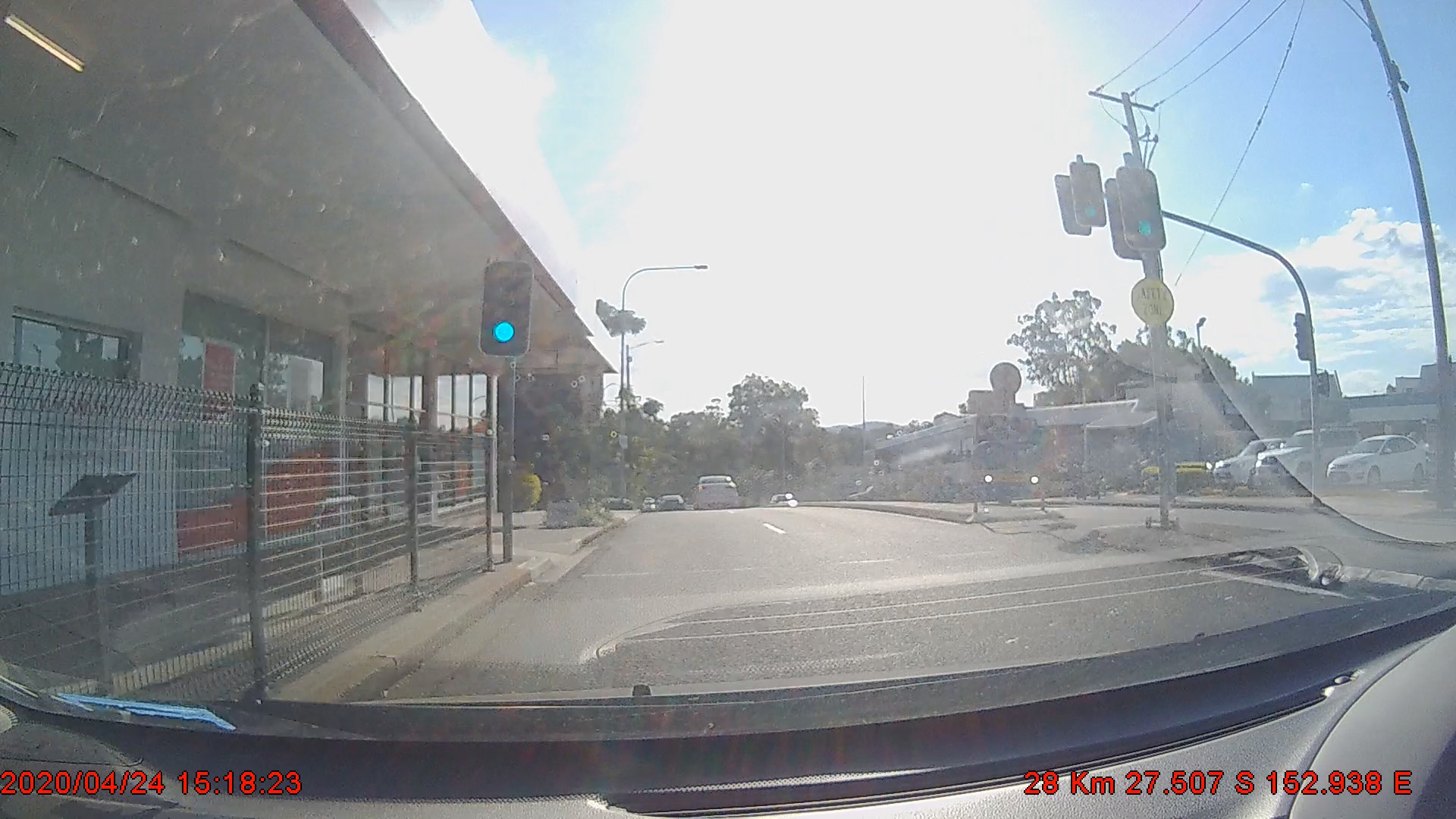}
      \includegraphics[width=0.22\linewidth]{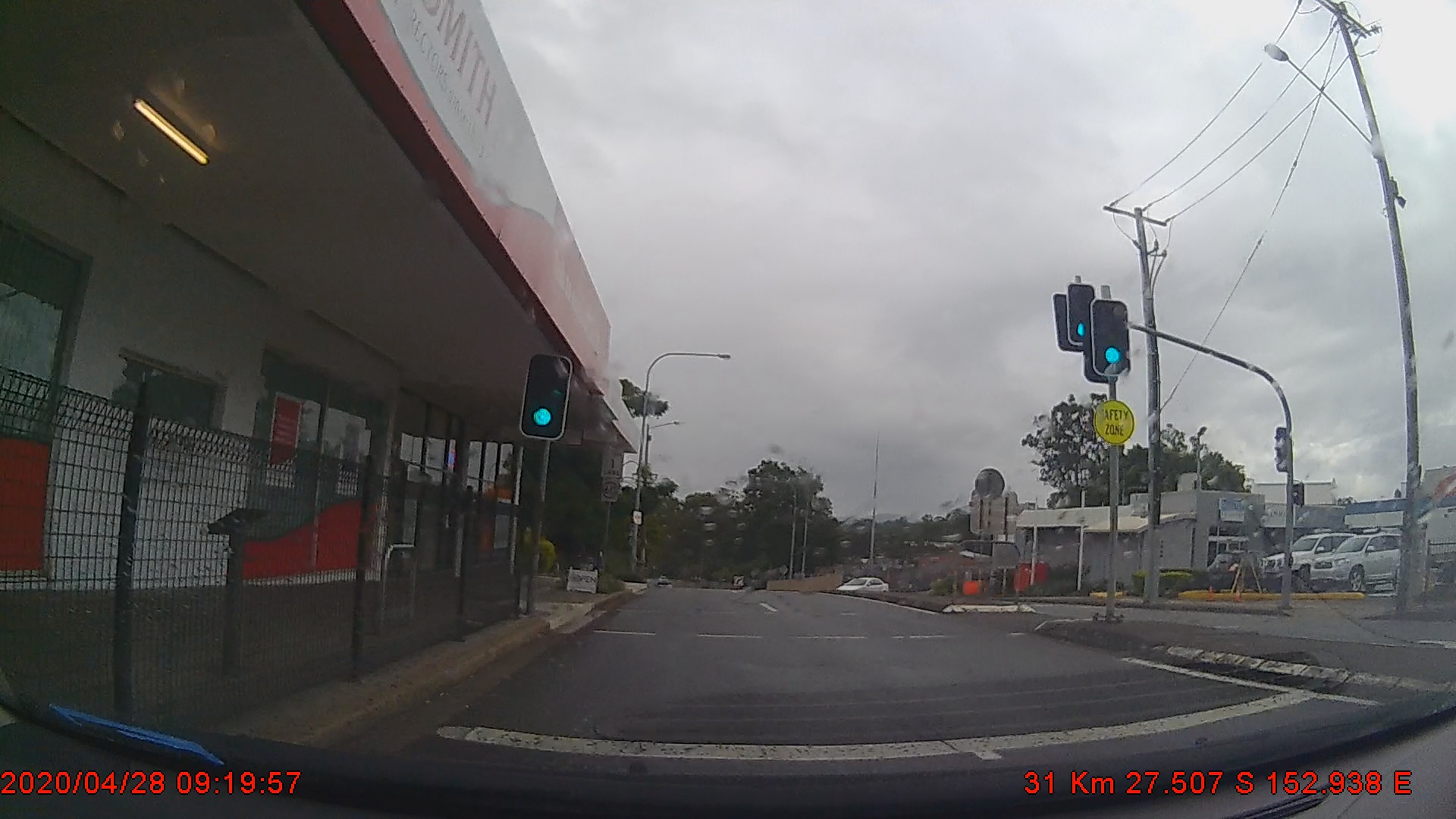}
  }\\[0.5cm]
  \makebox[.03\linewidth]{}
  \makebox[.48\linewidth]{Intersection}\hfil
  \makebox[.48\linewidth]{Avenue\ \ \ \ \ \ \ \ }\\[0.1cm]
  \rotatebox[origin=c]{90}{RGB frames}
  \raisebox{-0.45\height}{
      \includegraphics[width=0.22\linewidth]{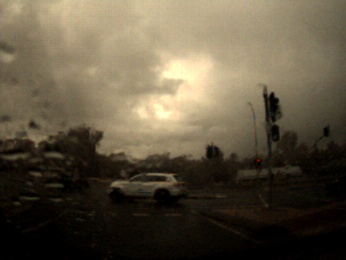}
      \includegraphics[width=0.22\linewidth]{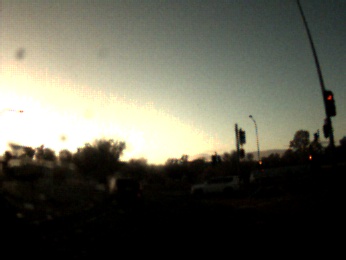}\hspace{0.01\linewidth}
      \includegraphics[width=0.22\linewidth]{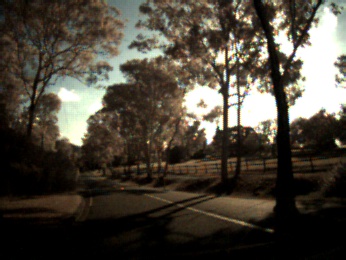}
      \includegraphics[width=0.22\linewidth]{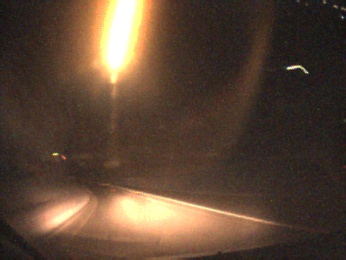}
  }\\[0.1cm]
  \rotatebox[origin=c]{90}{Reconstructed}
  \raisebox{-0.45\height}{
      \includegraphics[width=0.22\linewidth]{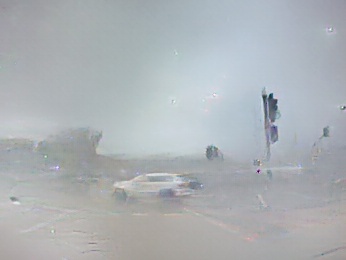}
      \includegraphics[width=0.22\linewidth]{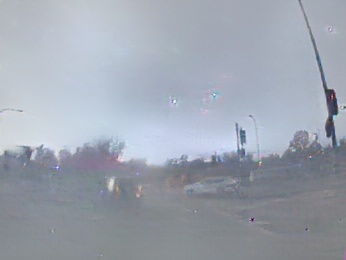}\hspace{0.01\linewidth}
      \includegraphics[width=0.22\linewidth]{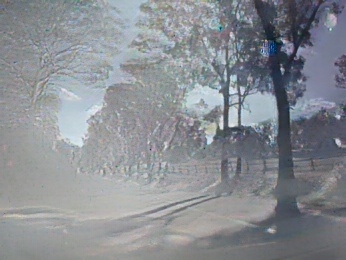}
      \includegraphics[width=0.22\linewidth]{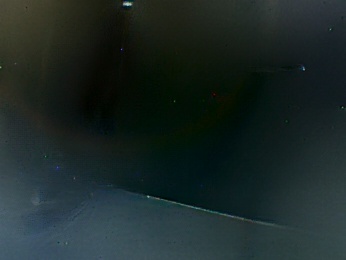}
  }\\[0.1cm]
  \rotatebox[origin=c]{90}{Consumer cam.}
  \raisebox{-0.45\height}{
      \includegraphics[width=0.22\linewidth]{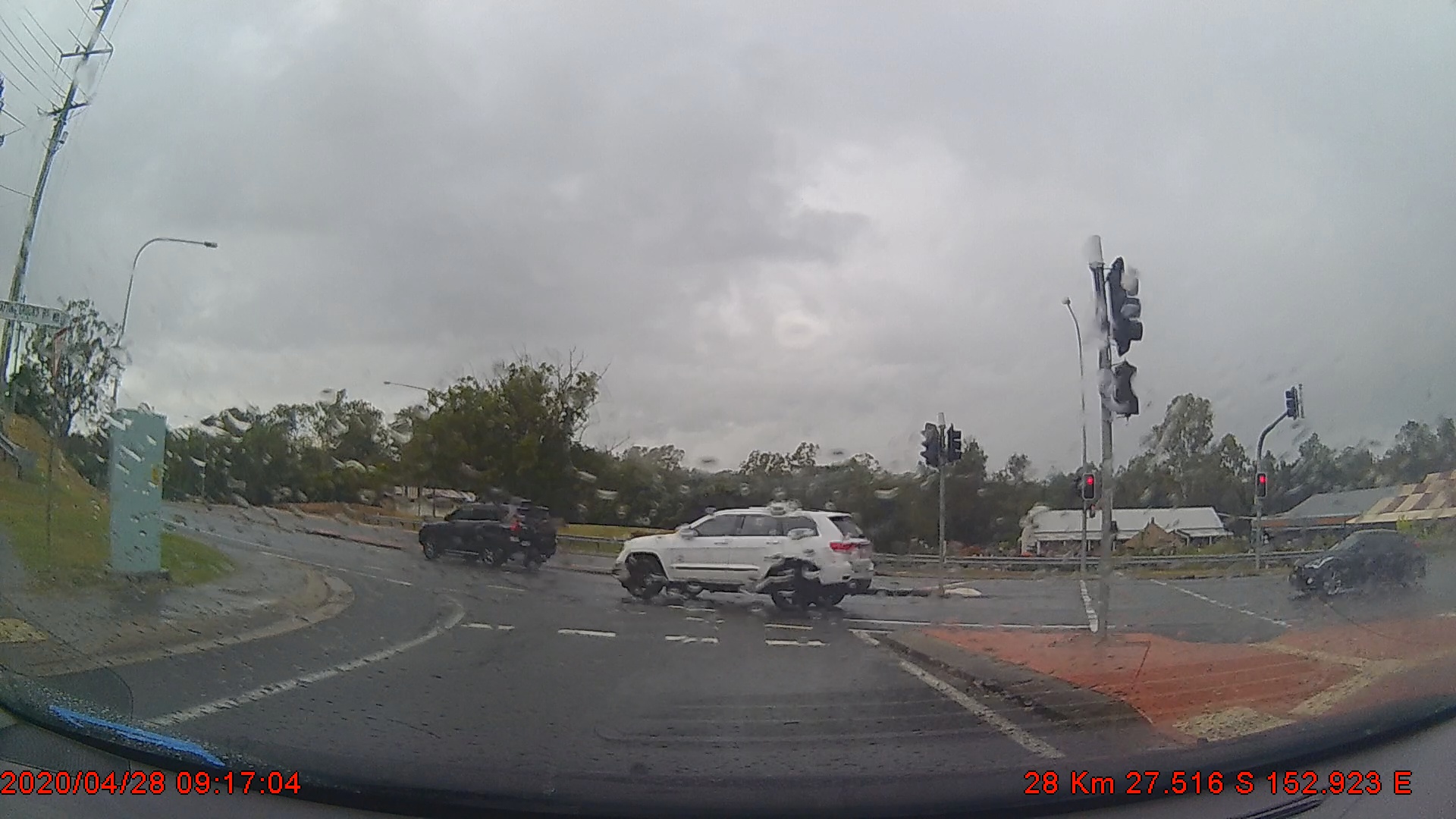}
      \includegraphics[width=0.22\linewidth]{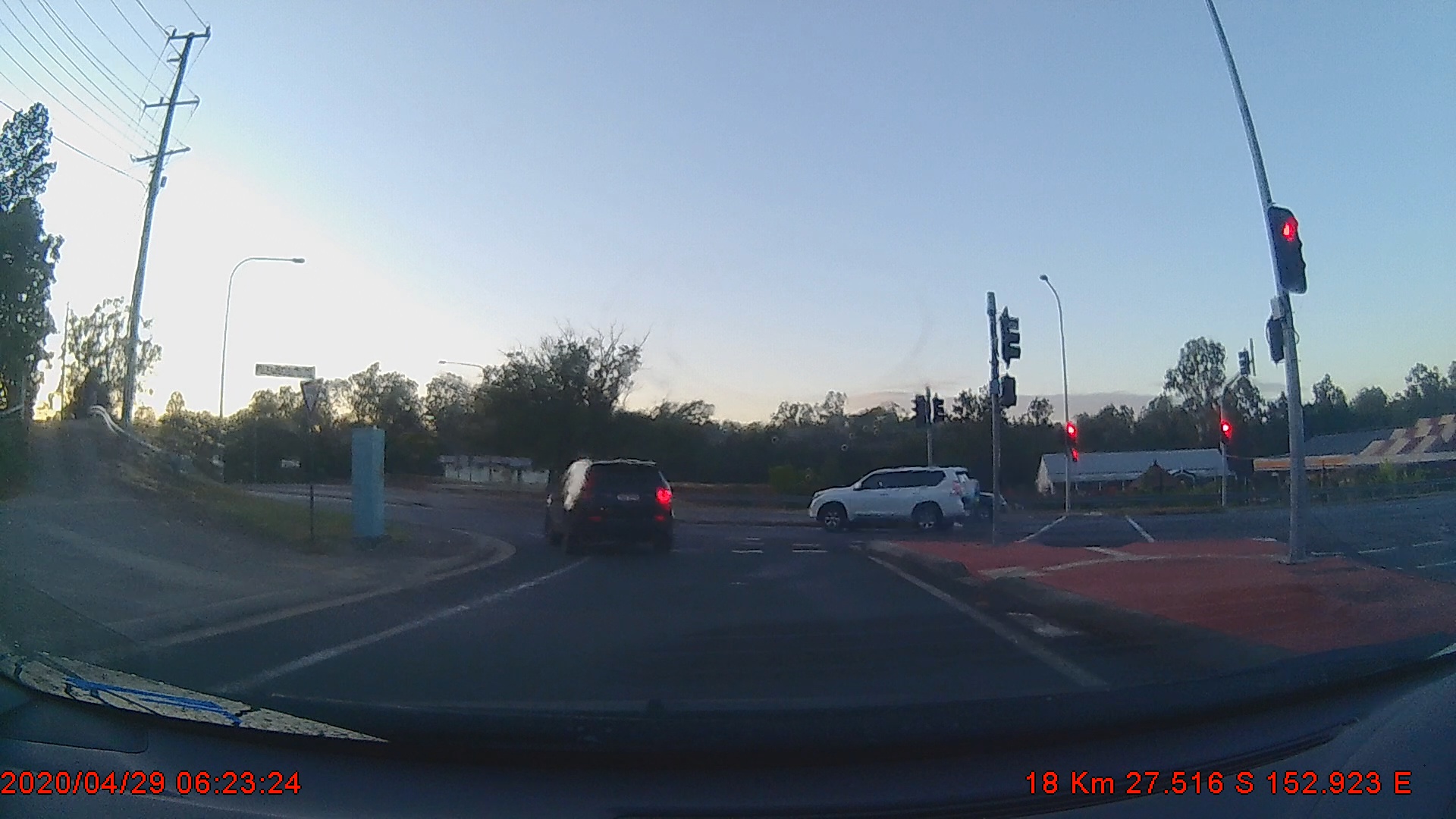}\hspace{0.01\linewidth}
      \includegraphics[width=0.22\linewidth]{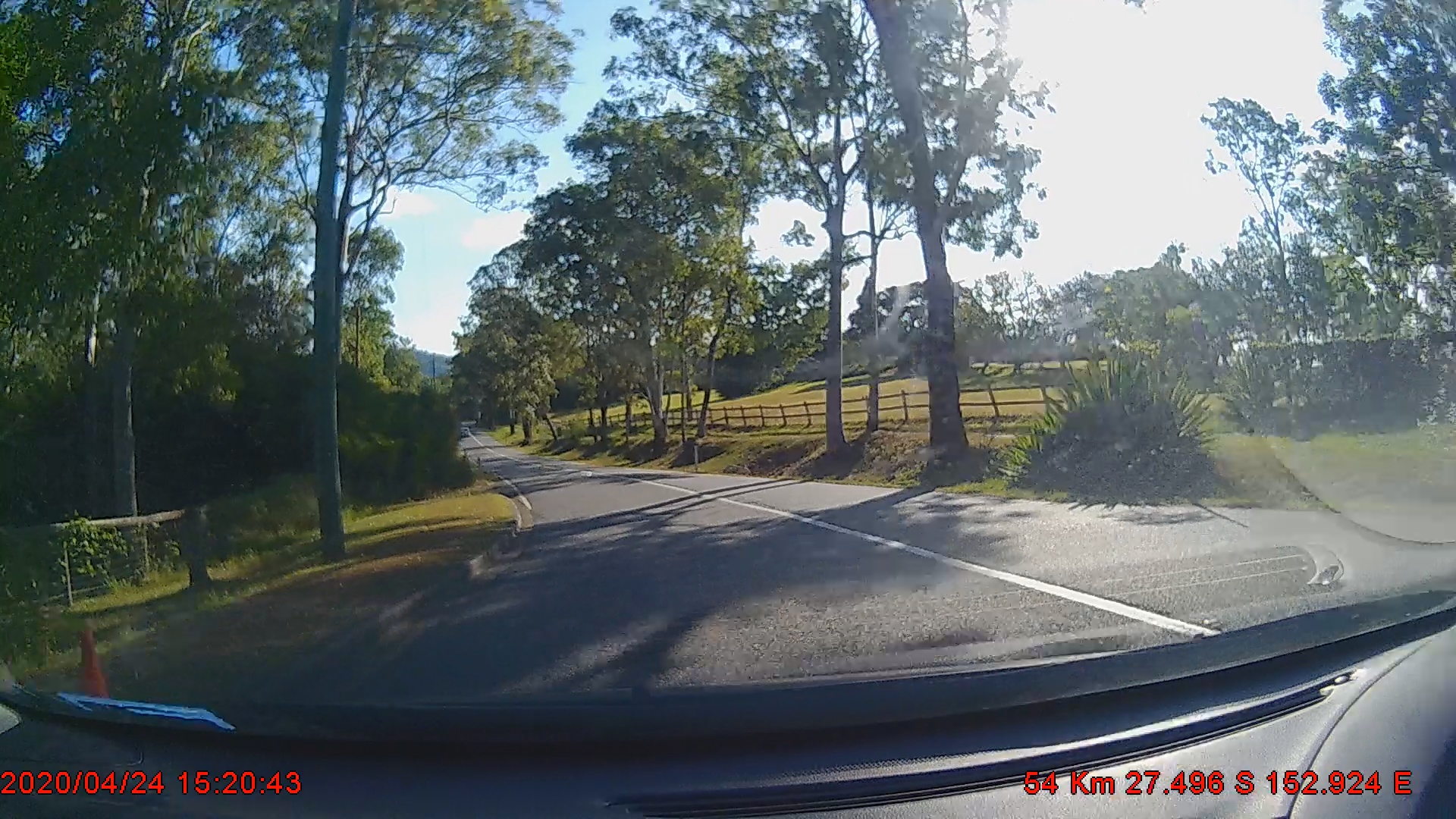}
      \includegraphics[width=0.22\linewidth]{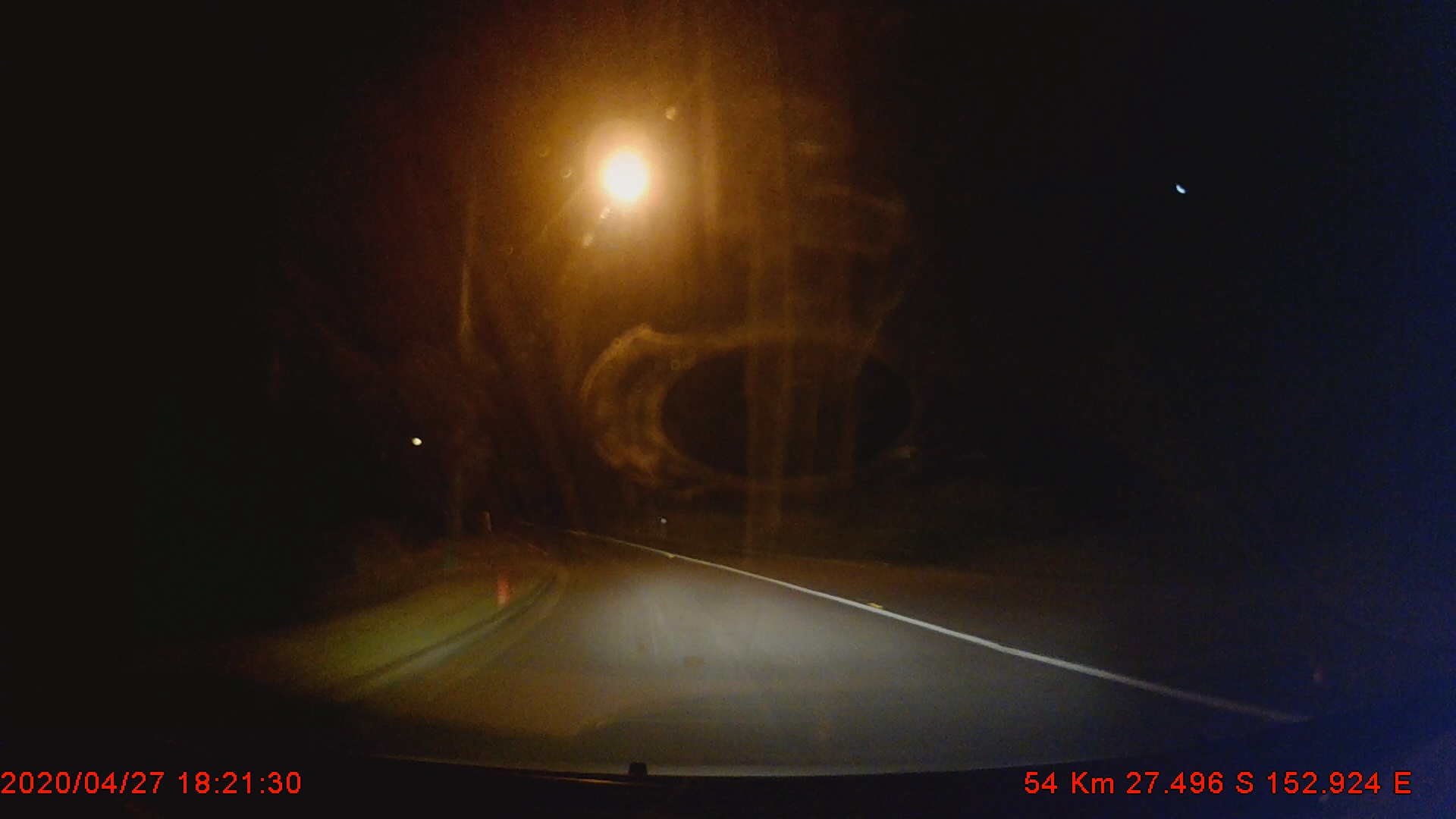}
  }
  \caption{Additional image pairs that are contained in the \protect\datasetname dataset (see also Fig.~2 of the main paper). In this figure we include images captured using a consumer camera.}%
  \label{fig:supp_example_images}%
\end{figure*}

In addition to the sample images presented in Fig.~2 of the main paper, Suppl.~Fig.~\ref{fig:supp_example_images} contains more examples of dataset images. We also add the relevant images captured by the consumer camera for reference.

\subsection{All combinations of traversals}
\begin{figure*}[t]
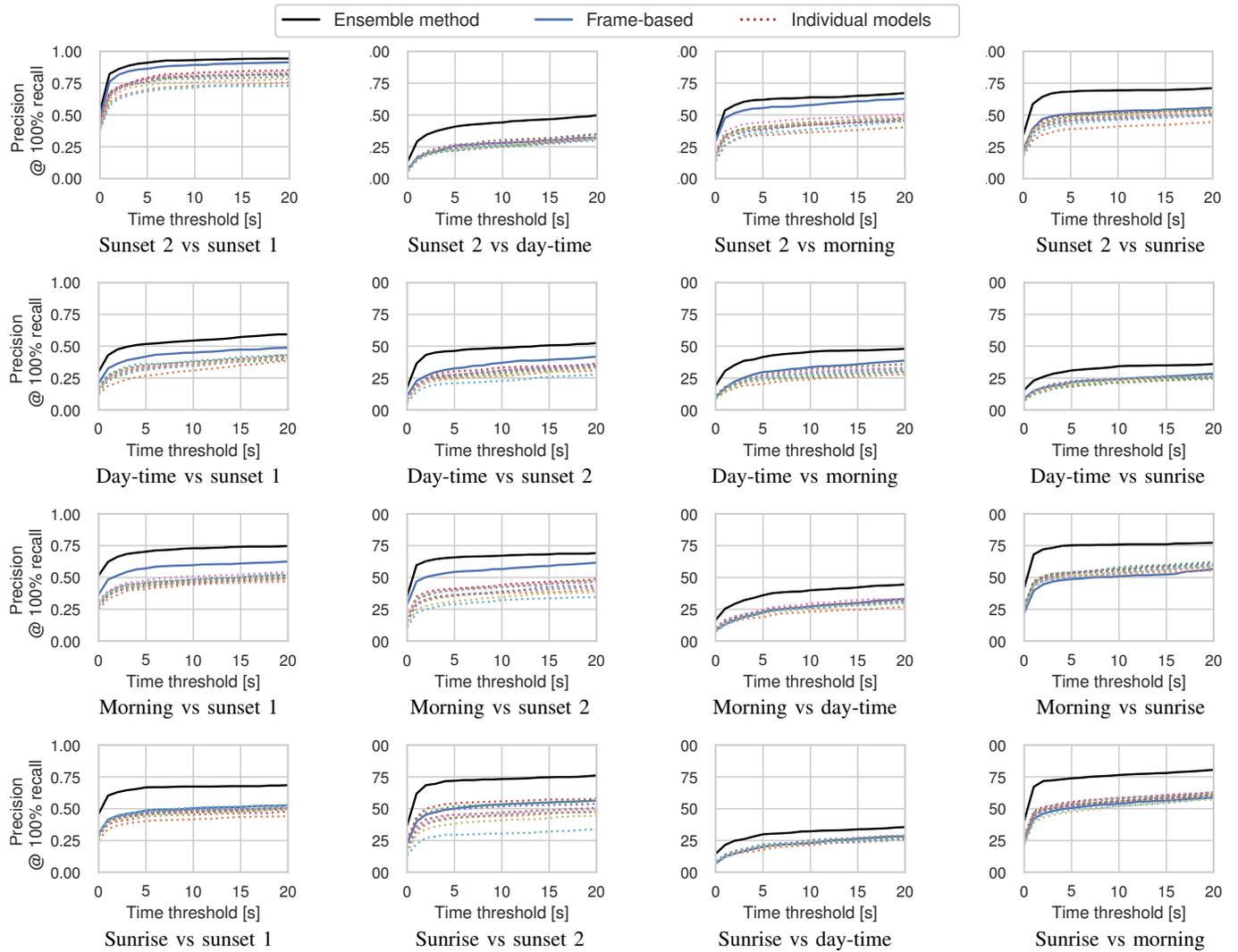

  \vspace{0.2cm}
  \makebox[\textwidth]{\resizebox{.6\textwidth}{!}{
    \inputpgf{figs}{recall_curve_legend.pgf}
  }}\\
  \resizebox{!}{.15\textwidth}{
    \begin{adjustbox}{clip,trim=0.5cm 0.28cm 0.42cm 0.25cm}
      \inputpgf{figs/recall_curves_supp}{recall_curve_netvlad_sunset2_vs_sunset1.pgf}
    \end{adjustbox}
  }
  \hfil
  \resizebox{!}{.15\textwidth}{
    \begin{adjustbox}{clip,trim=1.8cm 0.28cm 0.42cm 0.25cm}
      \inputpgf{figs/recall_curves_supp}{recall_curve_netvlad_sunset2_vs_daytime.pgf}
    \end{adjustbox}
  }
  \hfil
  \resizebox{!}{.15\textwidth}{
    \begin{adjustbox}{clip,trim=1.8cm 0.28cm 0.42cm 0.25cm}
      \inputpgf{figs/recall_curves_supp}{recall_curve_netvlad_sunset2_vs_morning.pgf}
    \end{adjustbox}
  }
  \hfil
  \resizebox{!}{.15\textwidth}{
    \begin{adjustbox}{clip,trim=1.8cm 0.28cm 0.42cm 0.25cm}
      \inputpgf{figs/recall_curves_supp}{recall_curve_netvlad_sunset2_vs_sunrise.pgf}
    \end{adjustbox}
  }
  \hfil
  \small
  \makebox[.025\textwidth]{}\makebox[.25\textwidth]{Sunset 2 vs sunset 1}\makebox[.26\textwidth]{Sunset 2 vs day-time}\makebox[.24\textwidth]{Sunset 2 vs morning}\makebox[.058\textwidth]{}\makebox[.15\textwidth][r]{Sunset 2 vs sunrise}\\[0.3cm]
  \resizebox{!}{.15\textwidth}{
    \begin{adjustbox}{clip,trim=0.5cm 0.28cm 0.42cm 0.25cm}
      \inputpgf{figs/recall_curves_supp}{recall_curve_netvlad_daytime_vs_sunset1.pgf}
    \end{adjustbox}
  }
  \hfil
  \resizebox{!}{.15\textwidth}{
    \begin{adjustbox}{clip,trim=1.8cm 0.28cm 0.42cm 0.25cm}
      \inputpgf{figs/recall_curves_supp}{recall_curve_netvlad_daytime_vs_sunset2.pgf}
    \end{adjustbox}
  }
  \hfil
  \resizebox{!}{.15\textwidth}{
    \begin{adjustbox}{clip,trim=1.8cm 0.28cm 0.42cm 0.25cm}
      \inputpgf{figs/recall_curves_supp}{recall_curve_netvlad_daytime_vs_morning.pgf}
    \end{adjustbox}
  }
  \hfil
  \resizebox{!}{.15\textwidth}{
    \begin{adjustbox}{clip,trim=1.8cm 0.28cm 0.42cm 0.25cm}
      \inputpgf{figs/recall_curves_supp}{recall_curve_netvlad_daytime_vs_sunrise.pgf}
    \end{adjustbox}
  }
  \hfil
  \small
  \makebox[.025\textwidth]{}\makebox[.25\textwidth]{Day-time vs sunset 1}\makebox[.26\textwidth]{Day-time vs sunset 2}\makebox[.24\textwidth]{Day-time vs morning}\makebox[.058\textwidth]{}\makebox[.15\textwidth][r]{Day-time vs sunrise}\\[0.3cm]
  \resizebox{!}{.15\textwidth}{
    \begin{adjustbox}{clip,trim=0.5cm 0.28cm 0.42cm 0.25cm}
      \inputpgf{figs/recall_curves_supp}{recall_curve_netvlad_morning_vs_sunset1.pgf}
    \end{adjustbox}
  }
  \hfil
  \resizebox{!}{.15\textwidth}{
    \begin{adjustbox}{clip,trim=1.8cm 0.28cm 0.42cm 0.25cm}
      \inputpgf{figs/recall_curves_supp}{recall_curve_netvlad_morning_vs_sunset2.pgf}
    \end{adjustbox}
  }
  \hfil
  \resizebox{!}{.15\textwidth}{
    \begin{adjustbox}{clip,trim=1.8cm 0.28cm 0.42cm 0.25cm}
      \inputpgf{figs/recall_curves_supp}{recall_curve_netvlad_morning_vs_daytime.pgf}
    \end{adjustbox}
  }
  \hfil
  \resizebox{!}{.15\textwidth}{
    \begin{adjustbox}{clip,trim=1.8cm 0.28cm 0.42cm 0.25cm}
      \inputpgf{figs/recall_curves_supp}{recall_curve_netvlad_morning_vs_sunrise.pgf}
    \end{adjustbox}
  }
  \hfil
  \small
  \makebox[.025\textwidth]{}\makebox[.25\textwidth]{Morning vs sunset 1}\makebox[.26\textwidth]{Morning vs sunset 2}\makebox[.24\textwidth]{Morning vs day-time}\makebox[.058\textwidth]{}\makebox[.15\textwidth][r]{Morning vs sunrise}\\[0.3cm]
  \resizebox{!}{.15\textwidth}{
    \begin{adjustbox}{clip,trim=0.5cm 0.28cm 0.42cm 0.25cm}
      \inputpgf{figs/recall_curves_supp}{recall_curve_netvlad_sunrise_vs_sunset1.pgf}
    \end{adjustbox}
  }
  \hfil
  \resizebox{!}{.15\textwidth}{
    \begin{adjustbox}{clip,trim=1.8cm 0.28cm 0.42cm 0.25cm}
      \inputpgf{figs/recall_curves_supp}{recall_curve_netvlad_sunrise_vs_sunset2.pgf}
    \end{adjustbox}
  }
  \hfil
  \resizebox{!}{.15\textwidth}{
    \begin{adjustbox}{clip,trim=1.8cm 0.28cm 0.42cm 0.25cm}
      \inputpgf{figs/recall_curves_supp}{recall_curve_netvlad_sunrise_vs_daytime.pgf}
    \end{adjustbox}
  }
  \hfil
  \resizebox{!}{.15\textwidth}{
    \begin{adjustbox}{clip,trim=1.8cm 0.28cm 0.42cm 0.25cm}
      \inputpgf{figs/recall_curves_supp}{recall_curve_netvlad_sunrise_vs_morning.pgf}
    \end{adjustbox}
  }
  \hfil
  \small
  \makebox[.025\textwidth]{}\makebox[.25\textwidth]{Sunrise vs sunset 1}\makebox[.26\textwidth]{Sunrise vs sunset 2}\makebox[.24\textwidth]{Sunrise vs day-time}\makebox[.058\textwidth]{}\makebox[.15\textwidth][r]{Sunrise vs morning}  
  
  \caption{Further performance evaluation on the remaining traverses of \datasetname. This figure supplements Fig.~5 in the main paper. All results were obtained using NetVLAD features.}%
  \label{fig:recall_varying_threshold_supp}%
\end{figure*}

\begin{figure}[t]
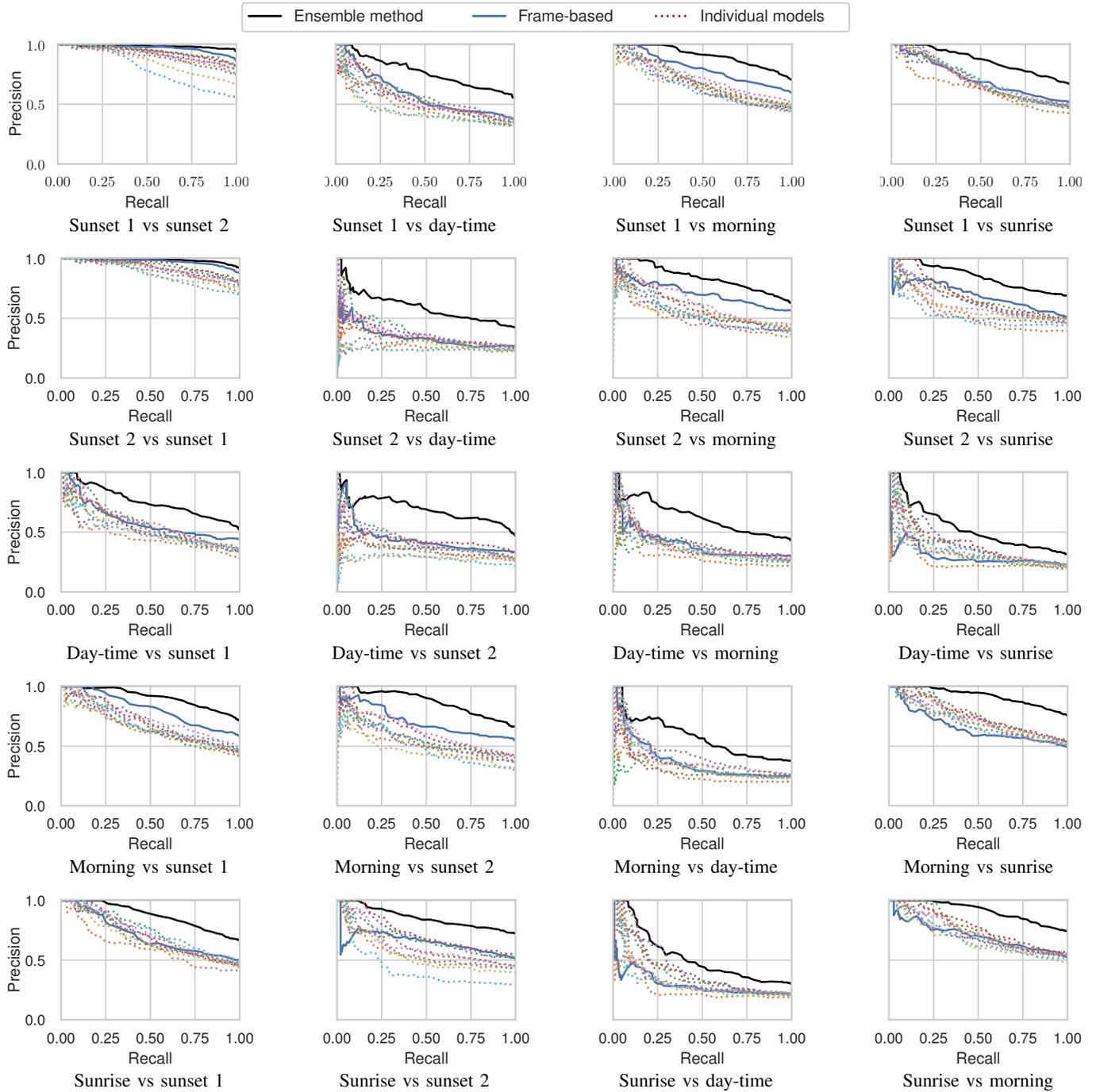

  \vspace{0.2cm}
  \makebox[\textwidth]{\resizebox{.6\textwidth}{!}{
    \inputpgf{figs}{recall_curve_legend.pgf}
  }}\\
  \resizebox{!}{.158\textwidth}{
    \begin{adjustbox}{clip,trim=0.55cm 0.3cm 0.42cm 0.25cm}
      \inputpgf{figs/pr_curves_supp}{pr_curve_netvlad_sunset1_vs_sunset2.pgf}
    \end{adjustbox}
  }
  \hfil
  \resizebox{!}{.158\textwidth}{
    \begin{adjustbox}{clip,trim=1.6cm 0.3cm 0.42cm 0.25cm}
      \inputpgf{figs/pr_curves_supp}{pr_curve_netvlad_sunset1_vs_daytime.pgf}
    \end{adjustbox}
  }
  \hfil
  \resizebox{!}{.158\textwidth}{
    \begin{adjustbox}{clip,trim=1.6cm 0.3cm 0.42cm 0.25cm}
      \inputpgf{figs/pr_curves_supp}{pr_curve_netvlad_sunset1_vs_morning.pgf}
    \end{adjustbox}
  }
  \hfil
  \resizebox{!}{.158\textwidth}{
    \begin{adjustbox}{clip,trim=1.6cm 0.3cm 0.42cm 0.25cm}
      \inputpgf{figs/pr_curves_supp}{pr_curve_netvlad_sunset1_vs_sunrise.pgf}
    \end{adjustbox}
  }
  \hfil
  \small
  \makebox[.01\textwidth]{}\makebox[.25\textwidth]{Sunset 1 vs sunset 2}\makebox[.24\textwidth]{Sunset 1 vs day-time}\makebox[.28\textwidth]{Sunset 1 vs morning}\makebox[.058\textwidth]{}\makebox[.132\textwidth][r]{Sunset 1 vs sunrise}\\[0.3cm]
  \resizebox{!}{.158\textwidth}{
    \begin{adjustbox}{clip,trim=0.55cm 0.3cm 0.42cm 0.25cm}
      \inputpgf{figs/pr_curves_supp}{pr_curve_netvlad_sunset2_vs_sunset1.pgf}
    \end{adjustbox}
  }
  \hfil
  \resizebox{!}{.158\textwidth}{
    \begin{adjustbox}{clip,trim=1.6cm 0.3cm 0.42cm 0.25cm}
      \inputpgf{figs/pr_curves_supp}{pr_curve_netvlad_sunset2_vs_daytime.pgf}
    \end{adjustbox}
  }
  \hfil
  \resizebox{!}{.158\textwidth}{
    \begin{adjustbox}{clip,trim=1.6cm 0.3cm 0.42cm 0.25cm}
      \inputpgf{figs/pr_curves_supp}{pr_curve_netvlad_sunset2_vs_morning.pgf}
    \end{adjustbox}
  }
  \hfil
  \resizebox{!}{.158\textwidth}{
    \begin{adjustbox}{clip,trim=1.6cm 0.3cm 0.42cm 0.25cm}
      \inputpgf{figs/pr_curves_supp}{pr_curve_netvlad_sunset2_vs_sunrise.pgf}
    \end{adjustbox}
  }
  \hfil
  \small
  \makebox[.01\textwidth]{}\makebox[.25\textwidth]{Sunset 2 vs sunset 1}\makebox[.24\textwidth]{Sunset 2 vs day-time}\makebox[.28\textwidth]{Sunset 2 vs morning}\makebox[.058\textwidth]{}\makebox[.132\textwidth][r]{Sunset 2 vs sunrise}\\[0.3cm]
  \resizebox{!}{.158\textwidth}{
    \begin{adjustbox}{clip,trim=0.55cm 0.3cm 0.42cm 0.25cm}
      \inputpgf{figs/pr_curves_supp}{pr_curve_netvlad_daytime_vs_sunset1.pgf}
    \end{adjustbox}
  }
  \hfil
  \resizebox{!}{.158\textwidth}{
    \begin{adjustbox}{clip,trim=1.6cm 0.3cm 0.42cm 0.25cm}
      \inputpgf{figs/pr_curves_supp}{pr_curve_netvlad_daytime_vs_sunset2.pgf}
    \end{adjustbox}
  }
  \hfil
  \resizebox{!}{.158\textwidth}{
    \begin{adjustbox}{clip,trim=1.6cm 0.3cm 0.42cm 0.25cm}
      \inputpgf{figs/pr_curves_supp}{pr_curve_netvlad_daytime_vs_morning.pgf}
    \end{adjustbox}
  }
  \hfil
  \resizebox{!}{.158\textwidth}{
    \begin{adjustbox}{clip,trim=1.6cm 0.3cm 0.42cm 0.25cm}
      \inputpgf{figs/pr_curves_supp}{pr_curve_netvlad_daytime_vs_sunrise.pgf}
    \end{adjustbox}
  }
  \hfil
  \small
  \makebox[.01\textwidth]{}\makebox[.25\textwidth]{Day-time vs sunset 1}\makebox[.24\textwidth]{Day-time vs sunset 2}\makebox[.28\textwidth]{Day-time vs morning}\makebox[.058\textwidth]{}\makebox[.132\textwidth][r]{Day-time vs sunrise}\\[0.3cm]
  \resizebox{!}{.158\textwidth}{
    \begin{adjustbox}{clip,trim=0.55cm 0.3cm 0.42cm 0.25cm}
      \inputpgf{figs/pr_curves_supp}{pr_curve_netvlad_morning_vs_sunset1.pgf}
    \end{adjustbox}
  }
  \hfil
  \resizebox{!}{.158\textwidth}{
    \begin{adjustbox}{clip,trim=1.6cm 0.3cm 0.42cm 0.25cm}
      \inputpgf{figs/pr_curves_supp}{pr_curve_netvlad_morning_vs_sunset2.pgf}
    \end{adjustbox}
  }
  \hfil
  \resizebox{!}{.158\textwidth}{
    \begin{adjustbox}{clip,trim=1.6cm 0.3cm 0.42cm 0.25cm}
      \inputpgf{figs/pr_curves_supp}{pr_curve_netvlad_morning_vs_daytime.pgf}
    \end{adjustbox}
  }
  \hfil
  \resizebox{!}{.158\textwidth}{
    \begin{adjustbox}{clip,trim=1.6cm 0.3cm 0.42cm 0.25cm}
      \inputpgf{figs/pr_curves_supp}{pr_curve_netvlad_morning_vs_sunrise.pgf}
    \end{adjustbox}
  }
  \hfil
  \small
  \makebox[.01\textwidth]{}\makebox[.25\textwidth]{Morning vs sunset 1}\makebox[.24\textwidth]{Morning vs sunset 2}\makebox[.28\textwidth]{Morning vs day-time}\makebox[.058\textwidth]{}\makebox[.132\textwidth][r]{Morning vs sunrise}\\[0.3cm]
  \resizebox{!}{.158\textwidth}{
    \begin{adjustbox}{clip,trim=0.55cm 0.3cm 0.42cm 0.25cm}
      \inputpgf{figs/pr_curves_supp}{pr_curve_netvlad_sunrise_vs_sunset1.pgf}
    \end{adjustbox}
  }
  \hfil
  \resizebox{!}{.158\textwidth}{
    \begin{adjustbox}{clip,trim=1.6cm 0.3cm 0.42cm 0.25cm}
      \inputpgf{figs/pr_curves_supp}{pr_curve_netvlad_sunrise_vs_sunset2.pgf}
    \end{adjustbox}
  }
  \hfil
  \resizebox{!}{.158\textwidth}{
    \begin{adjustbox}{clip,trim=1.6cm 0.3cm 0.42cm 0.25cm}
      \inputpgf{figs/pr_curves_supp}{pr_curve_netvlad_sunrise_vs_daytime.pgf}
    \end{adjustbox}
  }
  \hfil
  \resizebox{!}{.158\textwidth}{
    \begin{adjustbox}{clip,trim=1.6cm 0.3cm 0.42cm 0.25cm}
      \inputpgf{figs/pr_curves_supp}{pr_curve_netvlad_sunrise_vs_morning.pgf}
    \end{adjustbox}
  }
  \hfil
  \small
  \makebox[.01\textwidth]{}\makebox[.25\textwidth]{Sunrise vs sunset 1}\makebox[.24\textwidth]{Sunrise vs sunset 2}\makebox[.28\textwidth]{Sunrise vs day-time}\makebox[.058\textwidth]{}\makebox[.132\textwidth][r]{Sunrise vs morning}
  \caption{Precision-recall curves for all traverse combinations in the \datasetname\ dataset. All results were obtained using NetVLAD features.}%
  \label{fig:pr_supp}%
\end{figure}

In the main paper, we present results where a traversal at sunset time (2020-04-21 17:03) is compared with the following four traversals: sunset (2020-04-22 17:24), day time (2020-04-24 15:12), morning (2020-04-28 09:14) and sunrise (2020-04-29 06:20). In this Section we present results for the remaining traversal combinations. We include results for precision at 100\% recall with varying localization threshold in Suppl.~Fig.~\ref{fig:recall_varying_threshold_supp}, and precision-recall curves in Suppl.~Fig.~\ref{fig:pr_supp}. The precision-recall curves are generated by varying a similarity threshold, whereby a match for a query is only retrieved when its cosine distance is less than this threshold. 

Our proposed ensemble method performs favorable in all evaluation scenarios. The evaluation is only performed on features obtained using NetVLAD~\cite{Arandjelovic2018} as NetVLAD significantly outperforms the sum of absolute differences (as used in SeqSLAM~\cite{Milford2012}) as shown in the main paper. 

\begin{figure}
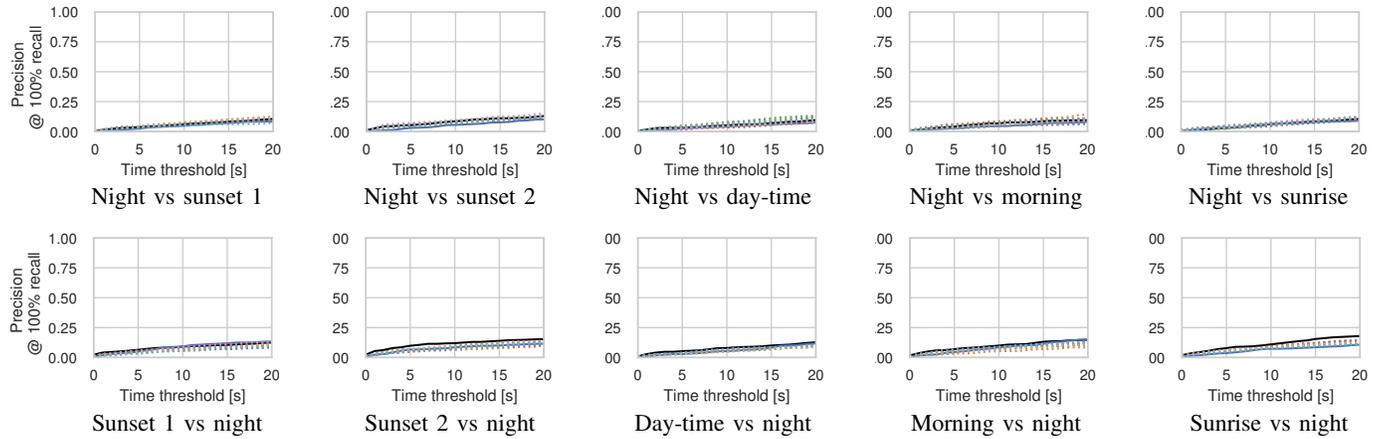

  \resizebox{!}{.126\textwidth}{
    \begin{adjustbox}{clip,trim=0.5cm 0.28cm 0.42cm 0.25cm}
      \inputpgf{figs/recall_curves_supp}{recall_curve_netvlad_night_vs_sunset1.pgf}
    \end{adjustbox}
  }
  \hfil
  \resizebox{!}{.126\textwidth}{
    \begin{adjustbox}{clip,trim=1.8cm 0.28cm 0.42cm 0.25cm}
      \inputpgf{figs/recall_curves_supp}{recall_curve_netvlad_night_vs_sunset2.pgf}
    \end{adjustbox}
  }
  \hfil
  \resizebox{!}{.126\textwidth}{
    \begin{adjustbox}{clip,trim=1.8cm 0.28cm 0.42cm 0.25cm}
      \inputpgf{figs/recall_curves_supp}{recall_curve_netvlad_night_vs_daytime.pgf}
    \end{adjustbox}
  }
  \hfil
  \resizebox{!}{.126\textwidth}{
    \begin{adjustbox}{clip,trim=1.8cm 0.28cm 0.42cm 0.25cm}
      \inputpgf{figs/recall_curves_supp}{recall_curve_netvlad_night_vs_morning.pgf}
    \end{adjustbox}
  }
  \hfil
  \resizebox{!}{.126\textwidth}{
    \begin{adjustbox}{clip,trim=1.8cm 0.28cm 0.42cm 0.25cm}
      \inputpgf{figs/recall_curves_supp}{recall_curve_netvlad_night_vs_sunrise.pgf}
    \end{adjustbox}
  }
  \hfil
  \small
  \makebox[.025\textwidth]{}\makebox[.2\textwidth]{Night vs sunset 1}\makebox[.2\textwidth]{Night vs sunset 2}\makebox[.2\textwidth]{Night vs day-time}\makebox[.2\textwidth]{Night vs morning}\makebox[.058\textwidth]{}\makebox[.1\textwidth][r]{Night vs sunrise}\\[0.3cm]
  \resizebox{!}{.126\textwidth}{
    \begin{adjustbox}{clip,trim=0.5cm 0.28cm 0.42cm 0.25cm}
      \inputpgf{figs/recall_curves_supp}{recall_curve_netvlad_sunset1_vs_night.pgf}
    \end{adjustbox}
  }
  \hfil
  \resizebox{!}{.126\textwidth}{
    \begin{adjustbox}{clip,trim=1.8cm 0.28cm 0.42cm 0.25cm}
      \inputpgf{figs/recall_curves_supp}{recall_curve_netvlad_sunset2_vs_night.pgf}
    \end{adjustbox}
  }
  \hfil
  \resizebox{!}{.126\textwidth}{
    \begin{adjustbox}{clip,trim=1.8cm 0.28cm 0.42cm 0.25cm}
      \inputpgf{figs/recall_curves_supp}{recall_curve_netvlad_daytime_vs_night.pgf}
    \end{adjustbox}
  }
  \hfil
  \resizebox{!}{.126\textwidth}{
    \begin{adjustbox}{clip,trim=1.8cm 0.28cm 0.42cm 0.25cm}
      \inputpgf{figs/recall_curves_supp}{recall_curve_netvlad_morning_vs_night.pgf}
    \end{adjustbox}
  }
  \hfil
  \resizebox{!}{.126\textwidth}{
    \begin{adjustbox}{clip,trim=1.8cm 0.28cm 0.42cm 0.25cm}
      \inputpgf{figs/recall_curves_supp}{recall_curve_netvlad_sunrise_vs_night.pgf}
    \end{adjustbox}
  }
  \hfil
  \small
  \makebox[.025\textwidth]{}\makebox[.2\textwidth]{Sunset 1 vs night}\makebox[.2\textwidth]{Sunset 2 vs night}\makebox[.2\textwidth]{Day-time vs night}\makebox[.2\textwidth]{Morning vs night}\makebox[.058\textwidth]{}\makebox[.1\textwidth][r]{Sunrise vs night}
    \caption{Performance evaluation on night traverses of \datasetname. As discussed in the main paper, none of the individual models performs better than random guessing; and thus the ensemble also leads to a random guess. All results were obtained using NetVLAD features.}
    \label{fig:recall_varying_threshold_supp_night}
\end{figure}

\begin{figure}
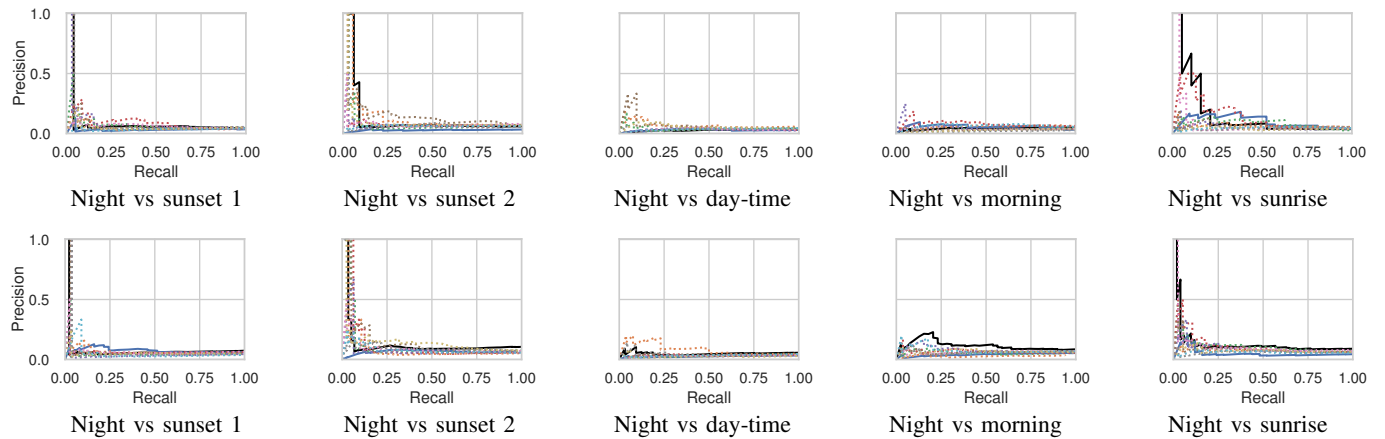

  \resizebox{!}{.126\textwidth}{
    \begin{adjustbox}{clip,trim=0.55cm 0.3cm 0.42cm 0.25cm}
      \inputpgf{figs/pr_curves_supp}{pr_curve_netvlad_night_vs_sunset1.pgf}
    \end{adjustbox}
  }
  \hfil
  \resizebox{!}{.126\textwidth}{
    \begin{adjustbox}{clip,trim=1.6cm 0.3cm 0.42cm 0.25cm}
      \inputpgf{figs/pr_curves_supp}{pr_curve_netvlad_night_vs_sunset2.pgf}
    \end{adjustbox}
  }
  \hfil
  \resizebox{!}{.126\textwidth}{
    \begin{adjustbox}{clip,trim=1.6cm 0.3cm 0.42cm 0.25cm}
      \inputpgf{figs/pr_curves_supp}{pr_curve_netvlad_night_vs_daytime.pgf}
    \end{adjustbox}
  }
  \hfil
  \resizebox{!}{.126\textwidth}{
    \begin{adjustbox}{clip,trim=1.6cm 0.3cm 0.42cm 0.25cm}
      \inputpgf{figs/pr_curves_supp}{pr_curve_netvlad_night_vs_morning.pgf}
    \end{adjustbox}
  }
  \hfil
  \resizebox{!}{.126\textwidth}{
    \begin{adjustbox}{clip,trim=1.6cm 0.3cm 0.42cm 0.25cm}
      \inputpgf{figs/pr_curves_supp}{pr_curve_netvlad_night_vs_sunrise.pgf}
    \end{adjustbox}
  }
  \hfil
  \small
  \makebox[.01\textwidth]{}\makebox[.2\textwidth]{Night vs sunset 1}\makebox[.2\textwidth]{Night vs sunset 2}\makebox[.2\textwidth]{Night vs day-time}\makebox[.2\textwidth]{Night vs morning}\makebox[.058\textwidth]{}\makebox[.1\textwidth][r]{Night vs sunrise}\\[0.3cm]
  \resizebox{!}{.126\textwidth}{
    \begin{adjustbox}{clip,trim=0.55cm 0.3cm 0.42cm 0.25cm}
      \inputpgf{figs/pr_curves_supp}{pr_curve_netvlad_sunset1_vs_night.pgf}
    \end{adjustbox}
  }
  \hfil
  \resizebox{!}{.126\textwidth}{
    \begin{adjustbox}{clip,trim=1.6cm 0.3cm 0.42cm 0.25cm}
      \inputpgf{figs/pr_curves_supp}{pr_curve_netvlad_sunset2_vs_night.pgf}
    \end{adjustbox}
  }
  \hfil
  \resizebox{!}{.126\textwidth}{
    \begin{adjustbox}{clip,trim=1.6cm 0.3cm 0.42cm 0.25cm}
      \inputpgf{figs/pr_curves_supp}{pr_curve_netvlad_daytime_vs_night.pgf}
    \end{adjustbox}
  }
  \hfil
  \resizebox{!}{.126\textwidth}{
    \begin{adjustbox}{clip,trim=1.6cm 0.3cm 0.42cm 0.25cm}
      \inputpgf{figs/pr_curves_supp}{pr_curve_netvlad_morning_vs_night.pgf}
    \end{adjustbox}
  }
  \hfil
  \resizebox{!}{.126\textwidth}{
    \begin{adjustbox}{clip,trim=1.6cm 0.3cm 0.42cm 0.25cm}
      \inputpgf{figs/pr_curves_supp}{pr_curve_netvlad_sunrise_vs_night.pgf}
    \end{adjustbox}
  }
  \hfil
  \small
  \makebox[.01\textwidth]{}\makebox[.2\textwidth]{Night vs sunset 1}\makebox[.2\textwidth]{Night vs sunset 2}\makebox[.2\textwidth]{Night vs day-time}\makebox[.2\textwidth]{Night vs morning}\makebox[.058\textwidth]{}\makebox[.1\textwidth][r]{Night vs sunrise}

  \caption{Precision-recall curves for combinations including night traverses. As discussed in the main paper, none of the individual models performs better than random guessing; and thus the ensemble also leads to a random guess. All results were obtained using NetVLAD features.}
    \label{fig:pr_supp_night}
\end{figure}

Furthermore, Suppl.~Figs.~\ref{fig:recall_varying_threshold_supp_night} and \ref{fig:pr_supp_night} include results for combinations including night traverses. As discussed in the main paper, none of the individual models performs better than random guessing for night traverses; and thus the ensemble also leads to a random guess.

\subsection{Comparison of distance matrices}
\begin{figure*}[t]
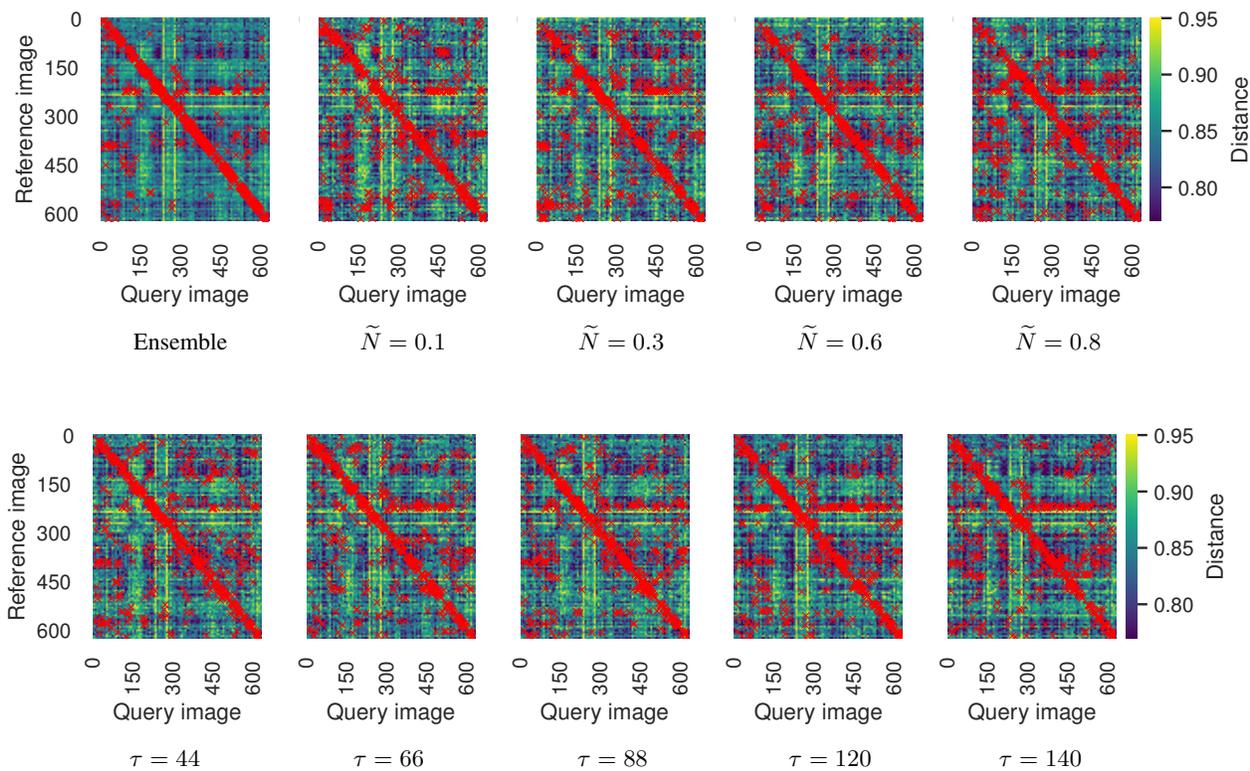

  \hfil
  \resizebox{!}{.22\textwidth}{
    \begin{adjustbox}{clip,trim=0.45cm 0.25cm 2.35cm 0.2cm}
      \inputpgf{figs/dist_matrices}{dist_matrix_netvlad_ensemble_frames.pgf}
    \end{adjustbox}
  }
  \resizebox{!}{.22\textwidth}{
    \begin{adjustbox}{clip,trim=1.75cm 0.25cm 2.35cm 0.2cm}
      \inputpgf{figs/dist_matrices}{dist_matrix_netvlad_fixed_num_events_0_1.pgf}
    \end{adjustbox}
  }
  \resizebox{!}{.22\textwidth}{
    \begin{adjustbox}{clip,trim=1.75cm 0.25cm 2.35cm 0.2cm}
      \inputpgf{figs/dist_matrices}{dist_matrix_netvlad_fixed_num_events_0_3.pgf}
    \end{adjustbox}
  }
  \resizebox{!}{.22\textwidth}{
    \begin{adjustbox}{clip,trim=1.75cm 0.25cm 2.35cm 0.2cm}
      \inputpgf{figs/dist_matrices}{dist_matrix_netvlad_fixed_num_events_0_6.pgf}
    \end{adjustbox}
  }
  \resizebox{!}{.22\textwidth}{
    \begin{adjustbox}{clip,trim=1.75cm 0.25cm 0.4cm 0.2cm}
      \inputpgf{figs/dist_matrices}{dist_matrix_netvlad_fixed_num_events_0_8.pgf}
    \end{adjustbox}
  }
  \hfil
  \\[0.2cm]
  \small
  \makebox[.2\textwidth][r]{Ensemble}\hfil\makebox[.06\textwidth][r]{$\widetilde{N}=0.1$}\hfil\makebox[.06\textwidth][r]{$\widetilde{N}=0.3$}\hfil\makebox[.06\textwidth][r]{$\widetilde{N}=0.6$}\hfil\makebox[.06\textwidth][r]{$\widetilde{N}=0.8$}\makebox[.06\textwidth][r]{}\\[1cm]
  \makebox[.0275\textwidth][r]{}
  \resizebox{!}{.22\textwidth}{
    \begin{adjustbox}{clip,trim=0.45cm 0.25cm 2.35cm 0.2cm}
      \inputpgf{figs/dist_matrices}{dist_matrix_netvlad_fixed_timespan_44.pgf}
    \end{adjustbox}
  }
  \resizebox{!}{.22\textwidth}{
    \begin{adjustbox}{clip,trim=1.75cm 0.25cm 2.35cm 0.2cm}
      \inputpgf{figs/dist_matrices}{dist_matrix_netvlad_fixed_timespan_66.pgf}
    \end{adjustbox}
  }
  \resizebox{!}{.22\textwidth}{
    \begin{adjustbox}{clip,trim=1.75cm 0.25cm 2.35cm 0.2cm}
      \inputpgf{figs/dist_matrices}{dist_matrix_netvlad_fixed_timespan_88.pgf}
    \end{adjustbox}
  }
  \resizebox{!}{.22\textwidth}{
    \begin{adjustbox}{clip,trim=1.75cm 0.25cm 2.35cm 0.2cm}
      \inputpgf{figs/dist_matrices}{dist_matrix_netvlad_fixed_timespan_120.pgf}
    \end{adjustbox}
  }
  \resizebox{!}{.22\textwidth}{
    \begin{adjustbox}{clip,trim=1.75cm 0.25cm 0.4cm 0.2cm}
      \inputpgf{figs/dist_matrices}{dist_matrix_netvlad_fixed_timespan_140.pgf}
    \end{adjustbox}
  }
  \\[0.2cm]
  \small
  \makebox[.18\textwidth][r]{$\tau=44$}\hfil\makebox[.06\textwidth][r]{$\tau=66$}\hfil\makebox[.06\textwidth][r]{$\tau=88$}\hfil\makebox[.06\textwidth][r]{$\tau=120$}\hfil\makebox[.05\textwidth][r]{$\tau=140$}\makebox[.07\textwidth][r]{}
  
  \caption{Comparison of distance matrices for the sunset vs morning comparison. Red dots indicate the best match for each query image. One can observe that the ensemble matrix is closest to an ideal place recognition algorithm, where all matches would lie on the diagonal.}%
  \label{fig:comparison_distance_matrices}%
\end{figure*}

In Suppl.~Fig.~\ref{fig:comparison_distance_matrices} we show the distance matrices obtained when comparing the sunset 1 traverse with the morning traverse for all individual ensemble members and the ensemble distance matrix. Red dots indicate the best match in the reference traverse. An ideal place recognition algorithm would lead to matches that lie on the diagonal. One can observe that there are fewer wrong matches (i.e.~matches off the diagonal) for the ensemble distance matrix when compared to all other individual distance matrices.

{
\hbadness 10000\relax
\bibliographystyle{IEEEtran}
\bibliography{references,references_ieeectrl}
}

\newpage

